\documentclass{bmvc2k}
\pdfoutput=1
\usepackage{bbding}

\usepackage{tikz}
\usepackage{comment}
\usepackage{amsmath,amssymb} 
\usepackage{color}

\usepackage{multirow}
\usepackage{booktabs}

\title{Deep Image Harmonization by Bridging the Reality Gap}

\addauthor{Junyan Cao$^\#$}{Joy_C1@sjtu.edu.cn}{1}
\addauthor{Wenyan Cong$^\#$}{wycong@utexas.edu}{1}
\addauthor{Li Niu$^\ast$}{ustcnewly@sjtu.edu.cn}{1}
\addauthor{Jianfu Zhang}{c.sis@sjtu.edu.cn}{1}
\addauthor{Liqing Zhang}{zhang-lq@cs.sjtu.edu.cn}{1}

\addinstitution{
 MoE Key Lab of Artificial Intelligence \\
 Shanghai Jiao Tong University \\
 Shanghai, China
}

\runninghead{Cao, Cong, Niu}{Deep Image Harmonization by Bridging the Reality Gap}

\def\etal{\emph{et al}\bmvaOneDot}

\begin{document}

\maketitle

\begin{abstract}
Image harmonization has been significantly advanced with large-scale harmonization dataset. However, the current way to build dataset is still labor-intensive, which adversely affects the extendability of dataset. To address this problem, we propose to construct rendered harmonization dataset with fewer human efforts to augment the existing real-world dataset. To leverage both real-world images and rendered images, we propose a cross-domain harmonization network to bridge the domain gap between two domains. Moreover, we also employ well-designed style classifiers and losses to facilitate cross-domain knowledge transfer. Extensive experiments demonstrate the potential of using rendered images for image harmonization and the effectiveness of our proposed network.
\end{abstract}

\section{Introduction}
\label{sec:intro}
Image composition \cite{niu2021making} combines foreground from one image and background from another image into one composite image. However, due to the appearance discrepancy caused by different capture conditions (\emph{e.g.} weather, season, time of the day) of foreground and background, the quality of composite image might be greatly degraded. Image harmonization aims to diminish the discrepancy by adjusting the appearance of the composite foreground according to the background. Recently, many deep learning based harmonization methods \cite{tsai2017deep,xiaodong2019improving,DoveNet2020,Hao2020bmcv,sofiiuk2021foreground,guo2021intrinsic,ling2021region} have achieved promising results.

Deep learning based methods require a large-scale training set that contains pairs of composite images and corresponding harmonized results. However, it is nontrivial and infeasible to collect abundant training pairs by manually harmonizing composite images. So inversely, \cite{tsai2017deep} and \cite{DoveNet2020} take real images as the ground-truth and adjust foreground appearance to generate synthetic pairs of composite images and ground-truth images. 
Though the above data acquisition is feasible in implementation, it still requires manual labor for foreground segmentation and unqualified composite image filtering~\cite{DoveNet2020}, which hinders the extendibility of harmonization dataset. For example, the foreground category of test composite images may be out of the scope of training set and we observe that cross-category harmonization suffers from a performance drop (see Section~\ref{sec:pre_exp}). However, extending the dataset to include novel foreground categories requires heavy human efforts. In the remainder of this paper, the category of a composite image means the category of its foreground to be harmonized. 

\begin{figure}[tp!]
\centering
\includegraphics[width=0.85\linewidth]{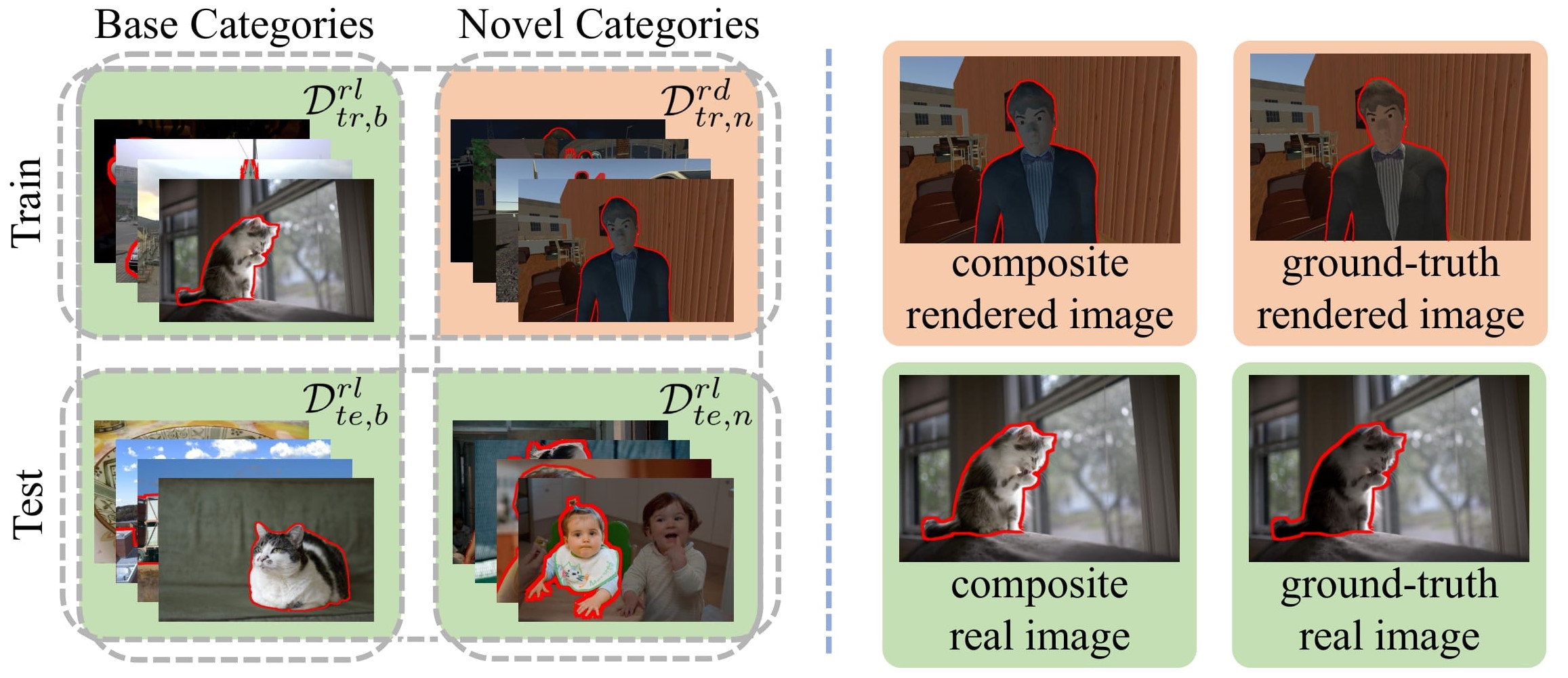}
\caption{Take ``human'' as an example novel category. Left sub-figure: illustration of training and test images from base categories and novel categories. Right sub-figure: image pairs from rendered image domain and real image domain. Images in orange (\emph{resp.}, green) are from rendered (\emph{resp.}, real) image domain. The foregrounds are outlined in red.}
\label{fig:traindata}
\end{figure}

In this work, we propose to construct image harmonization dataset using 3D rendering techniques, which can escape from heavy manual efforts and supplement existing real-world harmonization dataset. Specifically, we utilize 3D rendering software \emph{Unity3D} to generate rendered 2D scenes, and we leverage its plugin \emph{UniStorm} to generate a group of images with different capture conditions for each 2D scene. Since it is effortless to obtain accurate foreground masks in Unity3D, we could exchange the foregrounds within images from the same 2D scene, then obtain abundant pairs of composite rendered images and ground-truth rendered images.


Suppose that a real-world (``real'' for short) harmonization dataset has real images from base categories, we can enrich this dataset using rendered images from novel categories. Then, as in Figure~\ref{fig:traindata}, the image harmonization model could be trained on the combination of rendered images and real images, and tested on real images from both base and novel categories. This is a practical setting when our goal is enhancing the performance on novel categories which are out of the scope of real dataset. 

In this work, we contribute a Rendered Image Harmonization dataset \emph{RdHarmony}, which totally contains 6 categories (``human", ``bottle'', ``cake'', ``motorcycle'', ``cow'', and ``couch'').
In the main paper, we mainly take ``human'' as an example novel category because human harmonization is rather challenging and of paramount interest for many applications (\emph{e.g.}, augmented reality~\cite{tan2018and,weng2020misc}). 
The experiments using the other five categories as novel categories are left to the Supplementary.


However, models trained on rendered images (source domain) would suffer from a performance drop when applied to real images (target domain) (see Section~\ref{sec:pre_exp}), owning to considerably different data distributions between two domains, which is typically known as domain gap~\cite{torralba2011unbiased,segdomain2}. Hence, we propose a novel \textbf{C}ross-domain \textbf{harm}onization Network named \textbf{Charm}Net to align two domains when training with a combination of real images and rendered images. Our CharmNet has three stages. The \textit{domain-specific encoding stage} projects data from both domains to the same shared domain, where the projected features of two domains are further aligned with an adversarial loss. Then, the \textit{domain-invariant encoding-decoding stage} encourages information sharing between real and rendered images. The \textit{domain-specific decoding stage} projects the domain-invariant features back to different domains by enforcing the features to reconstruct the ground-truth images in each domain. 

Moreover, by naming different capture conditions as different ``styles'', we can easily obtain the style labels of rendered images using UniStorm. Therefore, we can explicitly exploit such information to facilitate cross-domain knowledge transfer. 
We assume that the input features before the second stage are inharmonious and the output features after the second stage are harmonious, so that the harmonization process is shared across two domains and more useful information can be transferred from rendered image domain to real image domain. For rendered images, their style labels before and after harmonization are readily available, which could be used to supervise style classification. For real images, their style labels before and after harmonization are unknown, but we assume that the style distribution should become more concentrated after harmonization because the foreground style is adapted to the background style. Hence, we propose a novel Style Aggregation (SA) loss to enforce the style distribution to be more concentrated after harmonization. 
To verify the effectiveness of our proposed network, we conduct comprehensive experiments on real dataset iHarmony4 and our contributed rendered dataset.

Our contributions are summarized as follows: 1) We investigate on the cross-domain and cross-category issues in image harmonization; 2) We have contributed and released the first large-scale rendered image harmonization dataset RdHarmony; 
3) We propose the first cross-domain image harmonization network CharmNet with novel network architecture and style aggregation loss;
4) Extensive experiments on real and rendered datasets demonstrate the potential of using rendered images for image harmonization.

\section{Related Works}

\subsection{Image Harmonization}\label{sec:RWIH}

Image harmonization has drawn increasing attention in the field of computer vision. Early works~\cite{colorharmonization,lalonde2007using,xue2012understanding,multi-scale,Pitie2005ndimensional,reinhard2001color,poisson,dragdroppaste,error-tolerant} mainly focused on manipulating the low-level image statistics, such as matching color distributions \cite{Pitie2005ndimensional,reinhard2001color}, applying gradient-domain compositing \cite{poisson,dragdroppaste,error-tolerant}, and mapping multi-scale statistics \cite{multi-scale}. Recently, deep learning based methods~\cite{tsai2017deep,DoveNet2020,Jiang_2021_ICCV,Guo_2021_ICCV,cong2021bargainnet} spring up rapidly. \cite{tsai2017deep,sofiiuk2021foreground} both leveraged auxiliary semantic features. \cite{DoveNet2020} released the first large-scale image harmonization dataset iHarmony4 and introduced a domain verification discriminator pulling close the foreground domain and background domain. \cite{xiaodong2019improving,Hao2020bmcv} explored various attention mechanisms. \cite{guo2021intrinsic,Guo_2021_ICCV} harmonized composite images by harmonizing reflectance and illumination separately. \cite{ling2021region} proposed to explicitly formulates background style and adaptively applied it to the foreground, which is extended in \cite{hang2022scs} by extracting the local background style relevant to the foreground. \cite{Cong_2022_CVPR} proposed to combine pixel-to-pixel transformation and color-to-color transformation in a unified framework. Different from existing methods only using real images, we are the first to use both real images and rendered images to train a cross-domain image harmonization model.

\subsection{Domain Adaptation}\label{sec:RWDA}

Domain adaptation strives to reduce the distribution mismatch and make the model trained on source domain generalize well to target domain. In our task, real images and rendered images are treated as two domains. Domain adaptation has been studied widely in classification \cite{classificationadaptation1,classficationadaptation2,clssegdomain},
detection~\cite{detectionadaptation,cyclegandetection,object3,object4}, segmentation~\cite{segdomain1,segdomain2,clssegdomain,advent}, person re-identification~\cite{reid1,reid2,reid3}, and pose estimation~\cite{posedomain3,posedomain4}. In the above applications, the label/output spaces (\emph{e.g.}, class label, segmentation mask) are the same between two domains. However, in image harmonization, the input and output spaces are the same in each domain but different across domains. Therefore, previous domain adaptation methods cannot be directly applied to our task. To the best of our knowledge, we are the first to address such a challenging task in the realm of domain adaptation.

Another possible domain adaptation approach is to translate images from the source domain to the target domain using Image-to-Image (I2I) translation methods~\cite{cyclegan,unit,pixelGAN,huang2018munit,starGan,lee2018diverse,lee2020drit++,liu2019few,park2020cut,Jung_2022_CVPR,mao2022continuous}, so that the translated images can be used for any downstream tasks. 
Nevertheless, during experiments, we observe that I2I translation methods have poor performance when translating rendered images to real images because they severely distort the illumination statistics. Therefore, I2I translation methods are ill-suited for our task.

\section{Dataset Construction}\label{sec:dataconstr}
We construct a rendered image harmonization dataset RdHarmony that contains rendered images with foregrounds of ``human'' category and other 5 object categories, \emph{i.e.}, ``bottle'', ``cake'', ``motorcycle'', ``cow'', and ``couch''.

The whole process can be divided into two steps. 
The first step is \emph{ground-truth rendered image generation}. We place various 3D characters of each novel category in different 3D scenes using Unity3D, and vary the camera viewpoints to generate 2D scenes. 
Meanwhile, we utilize UniStorm to control the weather and time. Note that UniStorm is a weather system plugin for Unity3D, which can simulate real-world natural phenomena by setting multiple attributes (\emph{e.g.}, light intensity, illumination color).
With UniStorm, each 2D scene could produce a group of images with different capture conditions (\emph{i.e.}, weather, time), which are referred to as different styles. In this work, we focus on natural lighting conditions because most images in iHarmony4~\cite{DoveNet2020} are captured in natural lighting conditions.
We select 3 representative weathers (Clear, Partly Cloudy, Cloudy) from UniStorm with pre-defined values. And we split time-of-the-day into 4 distinct phases (sunrise\&sunset, noon, night, other-times) by sampling every 5 minutes from 5:00 am to 21:00 pm. Based on the combinations of weather and time-of-the-day, we define 10 representative styles, including the night style as well as styles of Clear/Partly Cloudy/Cloudy weather at sunrise\&sunset/noon/other-times. \emph{Note that each style is not a discrete style, but covers a range of illumination intensity and direction. }

The second step is \emph{composite rendered image generation}. For each 2D scene, we treat one 3D character as foreground and obtain the foreground mask effortlessly using Unity3D. 
By exchanging the foregrounds of ground-truth rendered images in the same 2D scene with each other, we can produce composite rendered images with mixed styles. We define the style label of the ground-truth rendered image as a $10$-dim one-hot vector. Inspired by~\cite{cutmix}, we define a soft style label for the composite rendered image, which mixes the foreground and background style labels according to the foreground and background area ratio.
 
For ``human'' category, we generate 15,000 ground-truth rendered images with 1,500 different 2D scenes, and produce 135,000 rendered training pairs. For each object category (``bottle'', ``cake'', ``motorcycle'', ``cow'', and ``couch''), we create 9,000 rendered training pairs. Totally, our RdHarmony has 180,000 pairs of composite rendered images and ground-truth rendered images.
More details about RdHarmony are left to the Supplementary.

\begin{figure*}[tp!]
\centering
\includegraphics[width=0.90\linewidth]{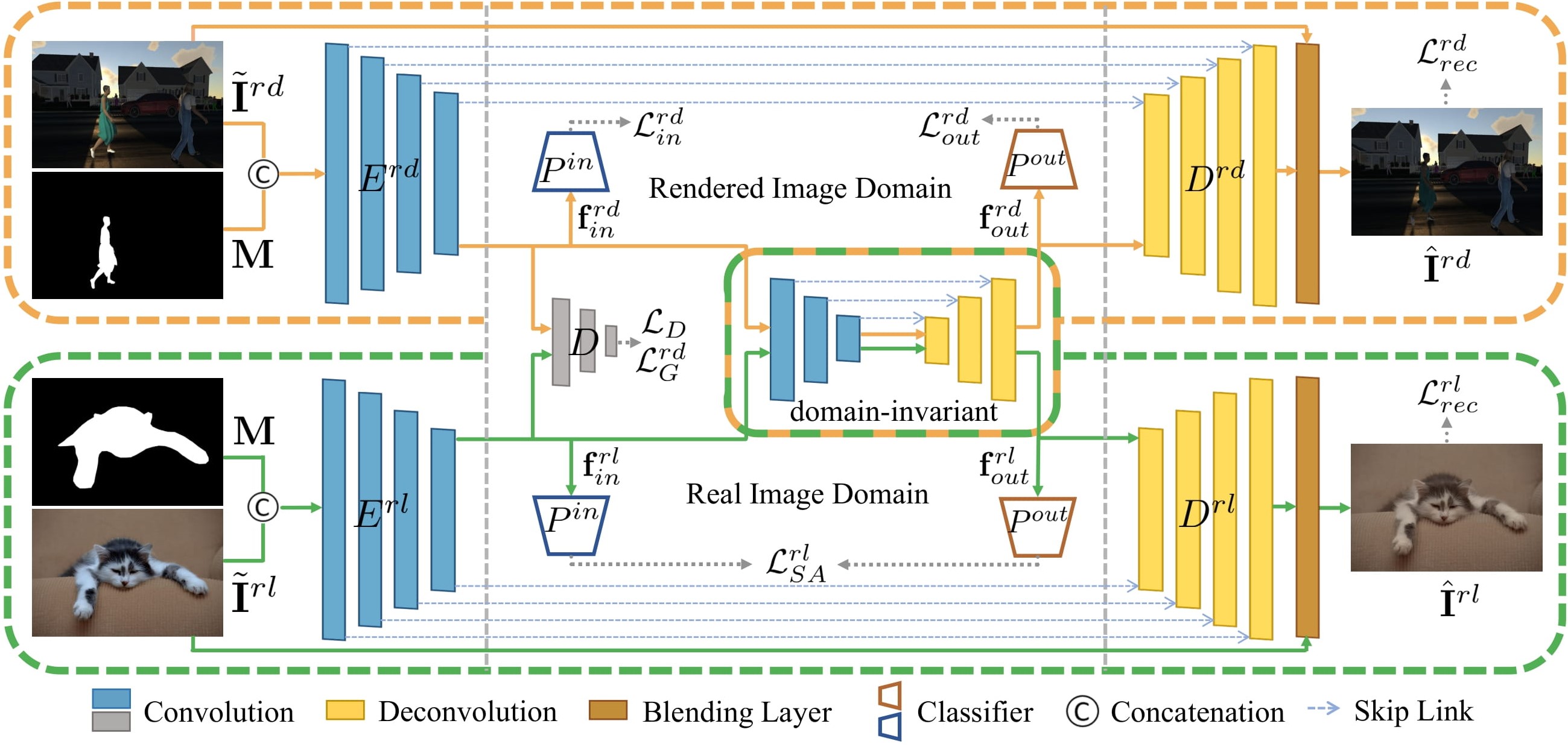}
\caption{The network architecture of CharmNet, which contains two three-stage generators, a domain discriminator $D$, and two style classifiers $P^{in}$ and $P^{out}$. The data flow of real (\emph{resp.}, rendered) images is marked with green (\emph{resp.}, orange) lines.}
\label{fig:network}
\end{figure*}

\section{Our Method}

In this work, we focus on training an image harmonization network using a combination of rendered image pairs and real image pairs.
In the training stage, we have access to rendered (rd) image pairs $\{\tilde{\mathbf{I}}^{rd}, \mathbf{I}^{rd}\}$ from novel categories and real (rl) image pairs $\{\tilde{\mathbf{I}}^{rl}, \mathbf{I}^{rl}\}$ from base categories, where $\tilde{\mathbf{I}}^{rd}$ (\emph{resp.}, $\tilde{ \mathbf{I}}^{rl}$) denotes the composite rendered (\emph{resp.}, real) image and $\mathbf{I}^{rd}$ (\emph{resp.}, $\mathbf{I}^{rl}$) denotes the corresponding ground-truth rendered (\emph{resp.}, real) image. We also have access to the binary foreground mask $\mathbf{M}$, which is not distinguished between domains. Given a composite rendered (\emph{resp.}, real) image $\tilde{\mathbf{I}}^{rd}$ (\emph{resp.}, $\tilde{\mathbf{I}}^{rl}$) and mask $\mathbf{M}$, the goal of image harmonization is to reconstruct $\mathbf{I}^{rd}$ (\emph{resp.}, $\mathbf{I}^{rl}$) with the harmonized rendered (\emph{resp.}, real) image $\hat{\mathbf{I}}^{rd}$ (\emph{resp.}, $\hat{\mathbf{I}}^{rl}$). In the testing stage, we aim to harmonize composite real images from both base and novel categories.


\subsection{Cross-domain Harmonization Network}\label{sec:charmnet}

As stated in Section \ref{sec:RWDA}, after harmonization, the composite rendered (\emph{resp.}, real) image is converted to harmonized rendered (\emph{resp.}, real) image. They share neither input space nor output space, so we design a three-stage network for cross-domain harmonization, including a domain-specific encoding stage, a domain-invariant encoding-decoding stage, and a domain-specific decoding stage. 

The pipeline of our network is shown in Figure~\ref{fig:network}. 
The generator $G$ adopts the improved backbone iDIH proposed in~\cite{sofiiuk2021foreground}, where an image blending layer is added to the UNet-like architecture. Since both input space and output space are different across two domains, we split the first (\emph{resp.}, last) 4 layers in the encoder (\emph{resp.}, decoder) into $E^{rd}$ (\emph{resp.}, $D^{rd}$) and $E^{rl}$ (\emph{resp.}, $D^{rl}$), where $E^{rd}$ (\emph{resp.}, $E^{rl}$) and $D^{rd}$ (\emph{resp.}, $D^{rl}$) are the domain-specific encoder and decoder for the rendered (\emph{resp.}, real) image domain. 

After the domain-specific encoding stage, we assume that images from different domains are projected into the same feature space. Inspired by~\cite{daadvbackprop,dann}, we employ a domain discriminator $D$ with adversarial loss~\cite{adda} to further align two domains, which is given by
\begin{eqnarray}\label{eq:gan}
\!\!\!\!&&\mathcal{L}_{D} = \mathbb{E}[\max(0, 1-D(\mathbf{f}^{rl}_{in}))] + \mathbb{E}[\max(0, 1+D(\mathbf{f}^{rd}_{in}))], \quad \mathcal{L}^{rd}_{G} = -\mathbb{E}[D(\mathbf{f}^{rd}_{in})],
\end{eqnarray}
where $\mathbf{f}^{rd}_{in}$ (\emph{resp.}, $\mathbf{f}^{rl}_{in}$) denotes the feature extracted by the domain-specific encoder $E^{rd}$ (\emph{resp.}, $E^{rl}$) in the rendered (\emph{resp.}, real) image domain. By playing a minimax game, the discriminator $D$ encourages the encoders $E^{rd}$ and $E^{rl}$ to project rendered and real images to the same feature space.


Then, through the domain-invariant encoder-decoder, the knowledge of harmonization is transferred from rendered image domain to real image domain. Finally, the third stage projects the domain-invariant features back to the input domain, by enforcing the harmonized output to approach the ground-truth image in each domain:
\begin{equation}\label{eq:l1}
    \begin{aligned}
\mathcal{L}^{rd}_{rec} = \|\hat{\mathbf{I}}^{rd}-\mathbf{I}^{rd}\|_1, \quad
\mathcal{L}^{rl}_{rec} = \|\hat{\mathbf{I}}^{rl}-\mathbf{I}^{rl}\|_1.
    \end{aligned}
\end{equation}

\subsection{Style Classifiers and Style Aggregation Loss}\label{sec:classifier}
When a composite rendered image passes through our three-stage network, we assume that the input feature before the second stage is inharmonious with mixed styles, while the output feature after the second stage is harmonious with a unified style. With this assumption, the main harmonization process is accomplished in the second stage, so that more useful knowledge can be transferred from rendered images to real images. Therefore, we employ two auxiliary style classifiers positioned before and after the second stage.

Because rendered images are associated with ground-truth style labels, we adopt standard cross-entropy classification losses for the inharmonious input feature $\mathbf{f}^{rd}_{in}$ and harmonious output feature $\mathbf{f}^{rd}_{out}$, which is given by
\begin{eqnarray}\label{eq:rd_all}
\mathcal{L}^{rd}_{in}=- \sum_{k=1}^{K} \tilde{y}_k^{rd} \log P_k^{in}\left(\mathbf{f}^{rd}_{in}\right), 
\quad
\mathcal{L}^{rd}_{out}=- \sum_{k=1}^{K} y_k^{rd} \log P_k^{out}\left(\mathbf{f}^{rd}_{out}\right),
\end{eqnarray}
\noindent where $K$ is the number of styles ($K=10$ as mentioned in Section~\ref{sec:dataconstr}). $\tilde y_{k}^{rd}$ (\emph{resp.}, $y_{k}^{rd}$) is the $k$-th entry of style label vector for the composite (\emph{resp.}, ground-truth) rendered image $\tilde{\mathbf{I}}^{rd}$ (\emph{resp.}, $\mathbf{I}^{rd}$). The style label vectors of rendered images have been introduced in Section~\ref{sec:dataconstr}.
$P^{in}$ (\emph{resp.}, $P^{out}$) is the style classifier before (\emph{resp.}, after) the second stage.
$P_k^{in}(\cdot)$ or $P_k^{out}(\cdot)$ is the $k$-th entry of predicted style distribution.

For real images without ground-truth style labels, standard classification loss cannot be used. Since the foreground style should be harmonized to the same as background style during harmonization, we design a novel style aggregation (SA) loss composed of two loss terms based on the following two assumptions. Firstly, $\mathbf{f}^{rl}_{out}$ is likely to have one style only if $\mathbf{f}^{rl}_{in}$ has this style. Thus, we adopt the weighted classification loss:
\begin{equation}\label{eq:rl_wc}
\begin{aligned}
\mathcal{L}^{rl}_{W}=&- \sum_{k=1}^{K} P_k^{in}\left(\mathbf{f}^{rl}_{in}\right) \log P_k^{out}\left(\mathbf{f}^{rl}_{out}\right).
\end{aligned}
\end{equation}
Intuitively, if $P_k^{in}\left(\mathbf{f}^{rl}_{in}\right)$ is small, $P_k^{out}\left(\mathbf{f}^{rl}_{out}\right)$ would also be penalized. Because rendered images and real images are projected to the same feature space after the first stage, $P^{in}$ and $P^{out}$ are shared across two domains.

Secondly, the style distribution of $\mathbf{f}^{rl}_{out}$ should be more concentrated than that of $\mathbf{f}^{rl}_{in}$. Since entropy depends on concentration and lower entropy implies higher concentration, we propose an entropy reduction loss to guarantee that  the style distribution becomes more concentrated after harmonization:
\begin{eqnarray}\label{eq:rl_er}
\mathcal{L}^{rl}_{ER}=max(0,m+\sum_{k=1}^{K} P_k^{in}\left(\mathbf{f}^{rl}_{in}\right) \log P_k^{in}\left(\mathbf{f}^{rl}_{in}\right)-  \sum_{k=1}^{K} P_k^{out}\left(\mathbf{f}^{rl}_{out}\right) \log P_k^{out}\left(\mathbf{f}^{rl}_{out}\right)),
\end{eqnarray}
where $m$ is a margin. We wrap up the above two loss terms as a style aggregation loss $\mathcal{L}^{rl}_{SA} = \mathcal{L}^{rl}_{W}+\mathcal{L}^{rl}_{ER}$. Therefore, the total loss function for training the generators is
\begin{eqnarray}\label{eq:loss}
\mathcal{L}_{G}= \mathcal{L}^{rd}_{rec}+\mathcal{L}^{rl}_{rec}+\lambda_{adv}\mathcal{L}^{rd}_{G}+\lambda_{sty}^{rd}(\mathcal{L}^{rd}_{in}+\mathcal{L}^{rd}_{out})+\lambda_{sty}^{rl}\mathcal{L}^{rl}_{SA},
\end{eqnarray}
where $\lambda_{adv}$, $\lambda_{sty}^{rd}$, and $\lambda_{sty}^{rl}$ are trade-off parameters. More implementation details can be found in the Supplementary.


\section{Experiments}

\subsection{Dataset}

\noindent\textbf{Rendered Image Dataset RdHarmony }contains 180,000 pairs of composite rendered images and ground-truth rendered images.
Note that we sample 65,000 pairs with ``human'' foregrounds from maximal 135,000 pairs to train the harmonization network, aiming to reduce data redundancy and improve training efficiency.

\noindent\textbf{Real Image Dataset iHarmony4~\cite{DoveNet2020}} contains 73,146 pairs of composite real images and ground-truth real images. In the main experiments, we treat ``human'' as an example novel category and other categories as base categories. The training set is split into novel training set (18,718 pairs) from ``human'' category and base training set (47,024 pairs) from other categories. Similarly, the test set is split into novel test set (1670 pairs) and base test set (5734 pairs). Experiments of the other 5 object categories could be found in the Supplementary.

Formally, we denote the rendered training set from novel category as $\mathcal{D}^{rd}_{tr,n}$ and the real training set from base (\emph{resp.}, novel) categories as $\mathcal{D}^{rl}_{tr,b}$ (\emph{resp.}, $\mathcal{D}^{rl}_{tr,n}$). We denote real test set from base (\emph{resp.}, novel) categories as $\mathcal{D}^{rl}_{te,b}$ (\emph{resp.}, $\mathcal{D}^{rl}_{te,n}$). By default, we merge $\mathcal{D}^{rd}_{tr,n}$ and $\mathcal{D}^{rl}_{tr,b}$ as the whole training set, and evaluate on both $\mathcal{D}^{rl}_{te,b}$ and $\mathcal{D}^{rl}_{te,n}$ (see Figure~\ref{fig:traindata}). 


%

\subsection{Preliminary Results} \label{sec:pre_exp}
We evaluate the performance of cross-category harmonization and cross-domain harmonization. We train iDIH on $\mathcal{D}^{rl}_{tr,n}$, $\mathcal{D}^{rl}_{tr,b}$ and $\mathcal{D}^{rd}_{tr,n}$ separately and test on $\mathcal{D}^{rl}_{te,n}$. As the size of $\mathcal{D}^{rl}_{tr,n}$ is $N=18718$, we sample $N$ training images from $\mathcal{D}^{rl}_{tr,b}$ (\emph{i.e.}, $\mathcal{D}^{rl}_{tr,b}(sub)$) and $\mathcal{D}^{rd}_{tr,n}$ (\emph{i.e.}, $\mathcal{D}^{rd}_{tr,n}(sub)$) for fair comparison.
From Table \ref{tab:gap_demon}, we observe that although cross-category harmonization could harmonize the composite images (column 3 \emph{v.s.} column 1), its performance is much worse than within-category harmonization (column 3 \emph{v.s.} column 2). We also observed that cross-domain harmonization hurts the performance badly, which demonstrates the large domain gap between rendered images and real images (column 4 \emph{v.s.} column 2).


In summary, without real training images from novel category $\mathcal{D}^{rl}_{tr,n}$, solely using $\mathcal{D}^{rl}_{tr,b}$ or using $\mathcal{D}^{rd}_{tr,n}$ will both lead to sub-optimal performance, which motivates us to train with a combination of $\mathcal{D}^{rl}_{tr,b}$ and $\mathcal{D}^{rd}_{tr,n}$.

\begin{table}
\scriptsize
\begin{center}
\setlength\tabcolsep{10pt}
\begin{tabular}{c|cccc}
\toprule \# & 1 & 2 & 3 & 4 \\
\hline Training data & - & $\mathcal{D}^{rl}_{tr,n}$ & $\mathcal{D}^{rl}_{tr,b}(sub)$ & $\mathcal{D}^{rd}_{tr,n}(sub)$ \\
\hline fMSE$\downarrow$ & 931.74 & 386.36 & 486.54 & 1100.86 \\
PSNR$\uparrow$ & 33.29 & 35.71 & 34.64 & 31.25 \\
\bottomrule

\end{tabular}
\end{center}
\caption{Results of models trained on $\mathcal{D}^{rl}_{tr,n}$, $\mathcal{D}^{rl}_{tr,b}(sub)$, and $\mathcal{D}^{rd}_{tr,n}(sub)$ and tested on $\mathcal{D}^{rl}_{te,n}$. ``-'' denotes the metrics directly tested on the composite real images.}
\label{tab:gap_demon}
\end{table}

\begin{table}[tb]
\scriptsize
\begin{center}
\begin{tabular}{c|c|c|ccc|ccc}
\toprule \multirow{2}*{\#
} & \multirow{2}*{Training data
} & \multirow{2}*{Method} & \multicolumn{3}{c|}{$\mathcal{D}^{rl}_{te,n}$} & \multicolumn{3}{c}{$\mathcal{D}^{rl}_{te,b}$}  \\
\cline{4-9}
~ & ~ & ~ & MSE$\downarrow$ & fMSE$\downarrow$ & PSNR$\uparrow$ & MSE$\downarrow$ & fMSE$\downarrow$ & PSNR$\uparrow$ \\
\hline
1 & - & Input Composite & 155.74 & 931.74 & 33.29 & 177.34 & 1505.92 & 31.15 \\ \hline 
2 & \multirow{5}*{$\mathcal{D}^{rl}_{tr,b}$} & DoveNet~\cite{DoveNet2020} & 69.06 & 458.15 & 35.01 & 65.87 & 674.57 & 33.94\\
3 & ~ & Hao \etal~\cite{Hao2020bmcv}& 52.42 & 407.26 & 35.43 & 58.21 & 582.79 & 34.54 \\
4 & ~ & iDIH~\cite{sofiiuk2021foreground} & 46.67 & 417.71 & 35.21 & 53.03 & 566.96 & 34.77 \\
5 & ~ & RainNet~\cite{ling2021region} & 58.40 & 525.36 & 34.97 & 59.54 & 701.07 & 34.44 \\
6 & ~ & Guo \etal~\cite{guo2021intrinsic} & 49.92 & 416.21 & 35.58 & 52.31 & 577.16 & 35.01 \\
\hline
7 & \multirow{5}*{$\mathcal{D}^{rd}_{tr,n}$ \& $\mathcal{D}^{rl}_{tr,b}$ } & two-stage training & 43.06 & 383.39 & 35.67 & 49.56 & 533.10 & 34.98\\
8 & ~ & dataset fusion & 38.31 & 368.50 & 35.64 & 44.15 & 485.94 & 35.27\\
\cline{3-9}
9 & ~ & UNIT~\cite{unit} & 55.09 & 458.72 & 34.74 & 56.21 & 607.14 & 34.38\\
10 & ~ & CycleGAN~\cite{cyclegan} & 51.82 & 474.78 & 34.64 & 56.30 & 592.33 & 34.52\\
11 & ~ & CUT~\cite{park2020cut} & 48.59 & 427.25 & 35.14 & 52.81 & 572.53 & 34.72\\
\cline{3-9} 
12 & ~ & CharmNet & \textbf{30.83} & \textbf{296.40} & \textbf{36.60} & \textbf{39.41} & \textbf{432.19} & \textbf{35.83}\\
\hline
13 & $\mathcal{D}^{rl}_{tr,n}$ \& $\mathcal{D}^{rl}_{tr,b}$  & iDIH~\cite{sofiiuk2021foreground} & 27.71 & 259.17 & 37.12 & 36.17 & 422.98 & 36.04\\
\bottomrule
\end{tabular}
\end{center}
\caption{Results of models trained on various training data and tested on $\mathcal{D}^{rl}_{te,n}$/$\mathcal{D}^{rl}_{te,b}$. ``-'' denotes metrics directly tested on composite real images. Note that row 13 using real training images $\mathcal{D}^{rl}_{tr,n}$ from novel category serves as the upper bound.}
\label{tab:baselines}
\end{table}

\subsection{Main Results} \label{sec:mainres}
As shown in Table~\ref{tab:baselines}, we divide baselines into two groups according to their used training data. The first group contains recent image harmonization baselines (row 2 to row 6) \cite{DoveNet2020,Hao2020bmcv,sofiiuk2021foreground,ling2021region,guo2021intrinsic} that are trained with only $\mathcal{D}^{rl}_{tr,b}$. Among them, iDIH~\cite{sofiiuk2021foreground} achieves competitive performance. Although the recent method \cite{Cong_2022_CVPR} achieves better performance, we opt for iDIH due to its simple network architecture. 
Thus, the rest of baselines (row 7 to row 12) and our method are all built upon iDIH. The second group contains domain adaptation models trained with both $\mathcal{D}^{rd}_{tr,n}$ and $\mathcal{D}^{rl}_{tr,b}$. Since there are no existing domain adaptation methods that could be directly applied to our task, we evaluate two straightforward methods, \emph{i.e.}, two-stage training and dataset fusion, as well as Image-to-Image (I2I) translation methods~\cite{unit,cyclegan,park2020cut} for comparison. Specifically, two-stage training means training with $\mathcal{D}^{rd}_{tr,n}$ and fine-tuning with $\mathcal{D}^{rl}_{tr,b}$. Dataset fusion means directly training with a mixture of $\mathcal{D}^{rd}_{tr,n}$ and $\mathcal{D}^{rl}_{tr,b}$. For I2I translation methods, we train UNIT \cite{unit}, CycleGAN \cite{cyclegan} 
, and CUT \cite{park2020cut} to translate from rendered image domain $\mathcal{D}^{rd}_{tr,n}$ to real image domain $\mathcal{D}^{rl}_{tr,b}$. Then, we use translated $\mathcal{D}^{rd}_{tr,n}$ to augment $\mathcal{D}^{rl}_{tr,b}$ as the new training set. 

The results of the second baseline group are also reported in Table~\ref{tab:baselines}. 
Two-stage training only brings minor performance gain (row 7 \emph{v.s.} row 4), probably because the statistics of real image domain are only considered during the second stage. 
The dataset fusion method improves the performance by a large margin (row 8 \emph{v.s.} row 4), which indicates that RdHarmony, though a rendered image dataset, is complementary to real image dataset by providing essential information of novel category.
Besides, the performances of UNIT, CycleGAN, and CUT are significantly degraded and even worse than iDIH (row 4), due to the low quality of translated images and the severely distorted illumination statistics. Our CharmNet outperforms all the baselines within the second group by a large margin. Apart from novel category, our method also significantly enhances the performance on base categories, which indicates that cross-domain cross-category knowledge transfer is also very useful. 

In addition, we report the result of iDIH trained with both $\mathcal{D}^{rl}_{tr,n}$ and $\mathcal{D}^{rl}_{tr,b}$ (row 13). This model serves as an upper bound for domain adaptation methods, assuming that there are abundant real training images from novel category. 
Despite the performance gap between row 12 and row 13, our method has shown great potential to bridge the gap between real images and rendered images. Since extending rendered image dataset costs significantly fewer manual efforts than extending real image dataset, we firmly believe that using both rendered images and real images for image harmonization is a promising research direction. 

\begin{figure*}[t]
\centering
\includegraphics[width=0.95\linewidth]{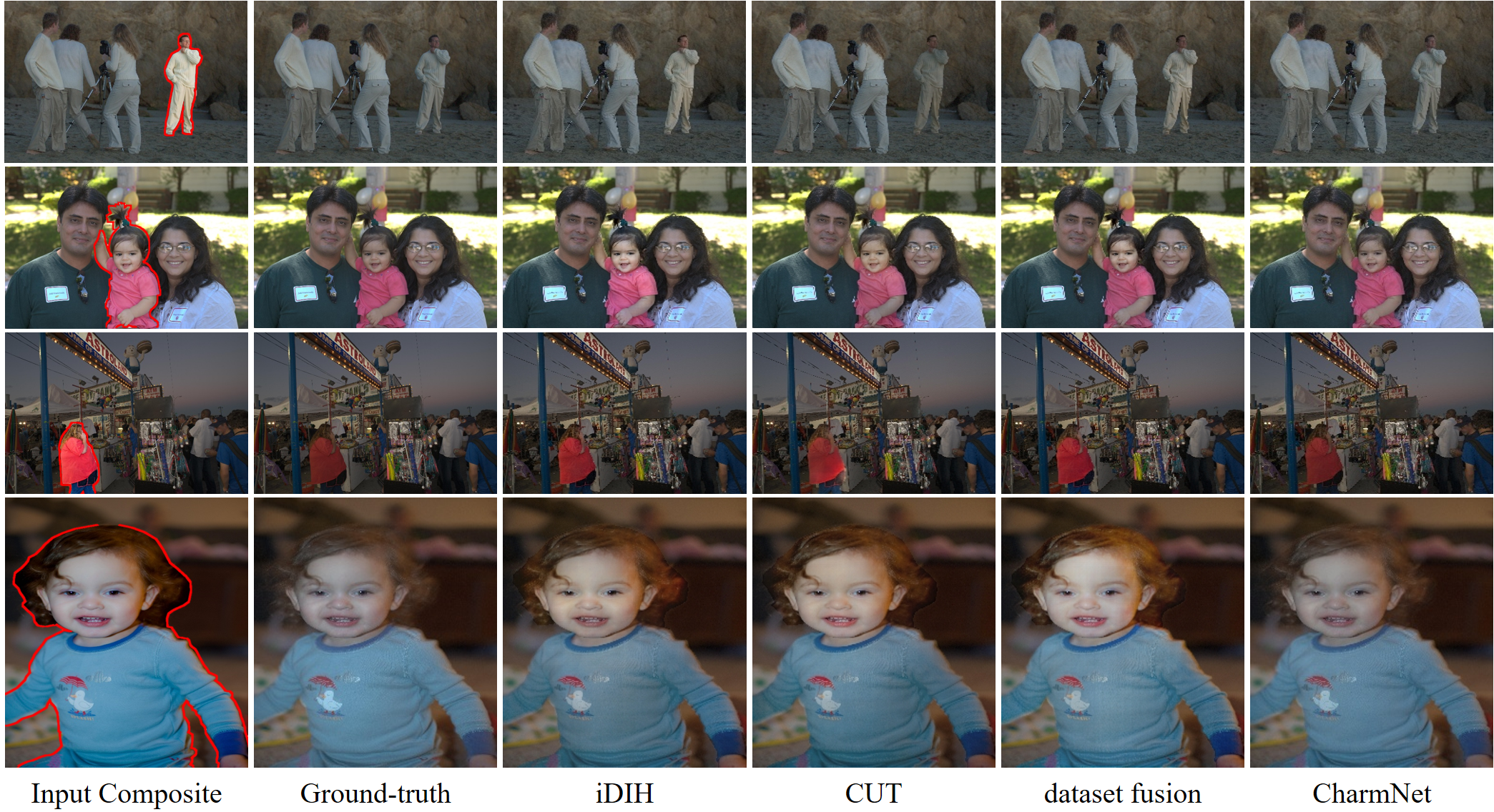}
\caption{Example results generated by different baselines and our method on $\mathcal{D}^{rl}_{te,n}$. From left to right, we show the input composite real image, ground-truth real image, as well as the harmonized images generated by iDIH (row 4 in Table~\ref{tab:baselines}), CUT, dataset fusion, and our CharmNet. The foregrounds are outlined in red.}
\label{fig:examples}
\end{figure*}

\subsection{Qualitative Analyses}\label{sec:qualitative}

Given a composite real image from $\mathcal{D}^{rl}_{te,n}$, the harmonized outputs generated by iDIH~\cite{sofiiuk2021foreground} (row 4 in Table~\ref{tab:baselines}), CUT~\cite{park2020cut}, dataset fusion, and our CharmNet are shown in Figure \ref{fig:examples}. Compared with other baselines, CharmNet could generate more favorable results with consistent foreground and background. Besides, CharmNet concentrates more on adjusting the foreground style to the background style without corrupting the appearance of human foreground, making the harmonized results visually closer to the ground-truth real images. In the Supplementary, we additionally provide example results on $\mathcal{D}^{rl}_{te,b}$ and compare the harmonization results of our method and upper bound (row 13 in Table~\ref{tab:baselines}).



\subsection{Ablation Studies} \label{sec:ablate}



We perform ablation studies to support the design of our CharmNet. First, we ablate the adversarial loss and style-related losses in Eqn.~\ref{eq:loss}.  The results demonstrate that each term is helpful and their combination achieves further improvement with mutual collaboration. 
Since the main challenge in our problem is to bridge the huge domain gap between rendered images and real images, so the adversarial loss contributes most to the performance gain. The two loss terms in style aggregation (SA) loss work together to meet our expectation, which leads to further improvement.
Next, we analyze the network design in terms of the number of unshared layers in the domain-specific encoding stage and domain-specific decoding stage (see Section~\ref{sec:charmnet}), which shows that it is necessary to use a suitable number of unshared layers. The detailed results are left to the Supplementary.

\subsection{Training Data Analyses}\label{sec:traindata}

We first explore how the diversity of RdHarmony dataset influences the harmonization performance by varying the number of 3D scenes and styles. Then with all 3D scenes and 10 styles, we explore the performance variance when using different numbers of rendered training images. We also explore the performance when using the whole $\mathcal{D}^{rd}_{tr,n}$ and partial real training images from novel category. In other words, we can jointly use rendered images and real images for the novel category. Specifically, we can use rendered training images to augment the real training images of novel category, leading to consistently better performance. When real training images are insufficient, the performance gain brought by rendered training images is more notable.
The detailed results are also left to the Supplementary.



\subsection{Evaluation on Real Composite Images}\label{sec:real_comp}

In real-world applications, there is no ground-truth for a real-world composite image with human foregrounds, so it is infeasible to evaluate the model performance quantitatively. Following \cite{tsai2017deep,DoveNet2020}, we select 46 images with human foregrounds from the real-world composite images~\cite{tsai2017deep} and conduct user study. Detailed results are left to the Supplementary.

\section{Conclusion}

In this paper, we have contributed a large-scale rendered image harmonization dataset RdHarmony, which simulates the process of capturing the same scene under different illumination conditions based on 3D rendering techniques. We have also presented a cross-domain image harmonization network CharmNet to bridge the gap between rendered images and real images. Our constructed dataset and proposed network have made a big step towards jointly using rendered images and real images for image harmonization.

\section*{Acknowledgement}

The work was supported by the Shanghai Municipal Science and Technology Major/Key Project, China (2021SHZDZX0102, 20511100300) and  National Natural Science Foundation of China (Grant No. 61902247).

\end{document}


\maketitle

In the Supplementary file, we will first introduce more details about our contributed dataset RdHarmony in Section~\ref{sec:dataconstr}, and introduce the implementation details in Section~\ref{sec:implement}. 
Then, we will ablate each component of the loss function, investigate the network design, and analyze the impact of different hyper-parameters in Section~\ref{sec:ablation}. 
We will explore the performance variance using different settings of training data in Section \ref{sec:supp_traindata}.
Besides, we will exhibit the harmonization results of different methods on real test set $\mathcal{D}^{rl}_{te,b}$ from base categories in Section~\ref{sec:res_real_base} and compare our method with the upper bound (row 13 in Table 2 in the main paper) in Section~\ref{sec:upper}. 
We will introduce more details of user study conducted on 46 real-world composite images with human foregrounds and show example harmonization results of different methods in Section~\ref{sec:real_comp}. 
Then, we will demonstrate the generalization of our CharmNet by conducting experiments on 5 more novel categories in Section~\ref{sec:gene_novel}.
Finally, we discuss the limitations in Section~\ref{sec:limit}. 

\section{Dataset Construction}\label{sec:dataconstr}

We construct rendered image harmonization dataset RdHarmony that contains rendered images with foregrounds of ``human'' and 5 object categories (``bottle'', ``cake'', ``motorcycle'', ``cow'', and ``couch'') from different super-categories.
Taking category ``human'' as an example, we first introduce the data generation process in Section~\ref{sec:realgen} and Section \ref{sec:compgen}. Then, we discuss the category extension in Section~\ref{sec:consider_ext} and illustrate the diversity in Section~\ref{sec:human_diversity}.


\subsection{Ground-truth Rendered Image Generation} \label{sec:realgen}
Generating a large number of rendered images with human foregrounds requires various 3D human characters placed in different 3D scenes. Considering the intra-category variance of ``human" category, we leverage the open-source software \emph{MakeHuman} to create diverse 3D human characters with distinct attributes including skeleton, body features (\emph{e.g.}, height, facial components), pose (\emph{e.g.}, walking, running), and clothes. 
Actually, due to the prevalence of 3D modelling, there are abundant available 3D models of ``human" category and other categories that could be used (see Section~\ref{sec:consider_ext}). 
Besides, we leverage the 3D game engine Unity3D and collect indoor and outdoor 3D scenes from Unity Asset Store and CG websites. Then, we import 3D human characters into the 3D scenes and vary the camera viewpoints to shoot various 2D scenes. In detail, we create 1,500 3D human characters, collect 30 3D scenes, and set 50 camera viewpoints for each 3D scene, leading to 1,500 2D scenes with unique foregrounds. 

As claimed in Section 3 in the main paper, we employ UniStorm to control the weather and time while producing 2D scenes. Based on the different capture conditions, we define 10 representative styles, including the night style as well as styles of Clear/Partly Cloudy/Cloudy weather at sunrise\&sunset/noon/other-times.
For each 2D scene, we randomly sample one rendered image from each style, resulting in a group of $10$ images. Given 1,500 2D scenes, we generate 15,000 rendered images with human foregrounds and different styles.

\begin{figure}[tp!]
\centering
    \includegraphics[width=0.85\linewidth]{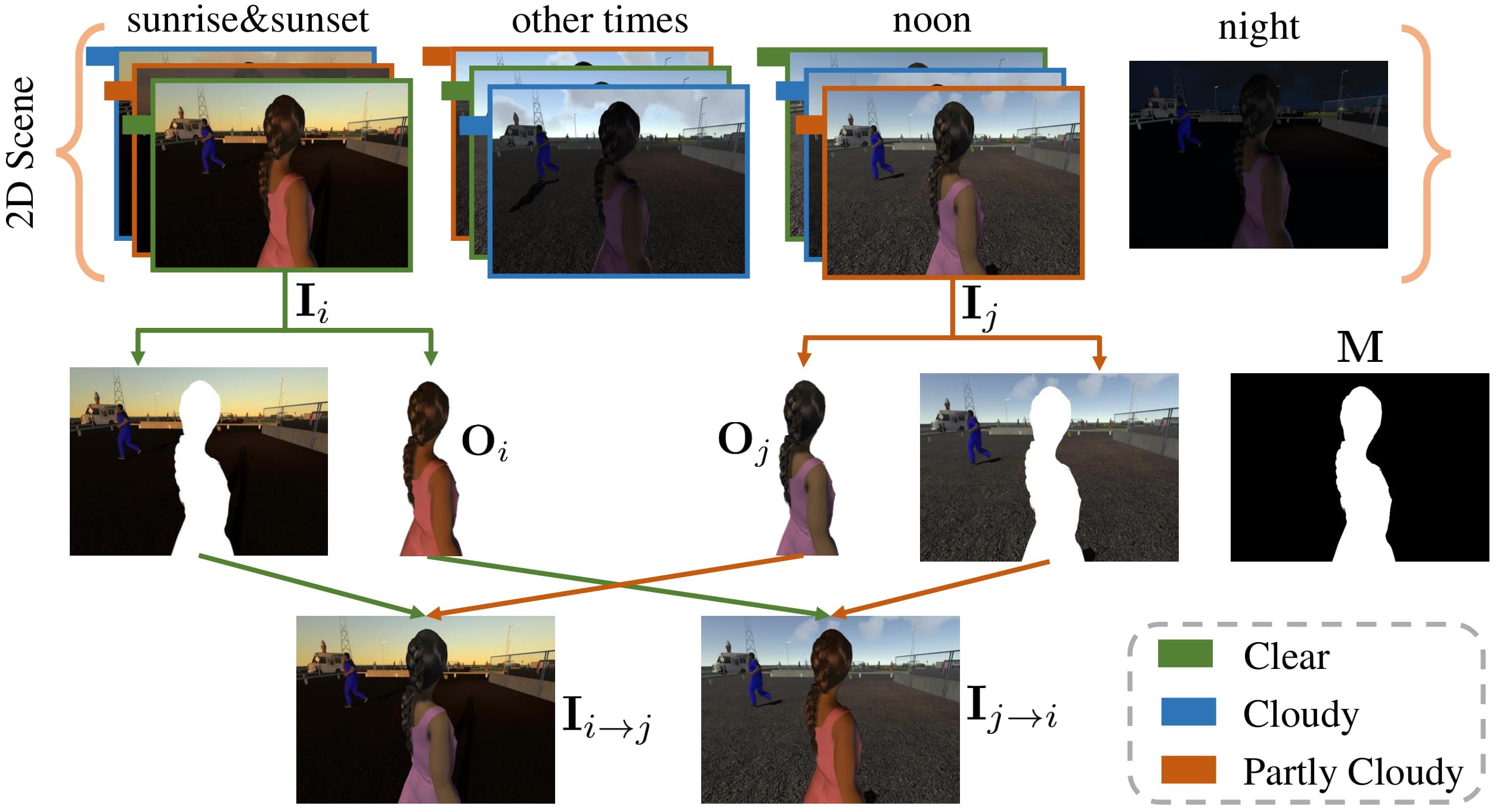}
\caption{The generation process of composite rendered images. Top row: one 2D scene with 10 different styles. Middle row: exchange the foregrounds between two rendered images with different styles. Bottom row: obtained composite rendered images.} 
\label{fig:dataset}
\end{figure}

\begin{figure*}[ht]
\centering
\includegraphics[width=0.99\linewidth]{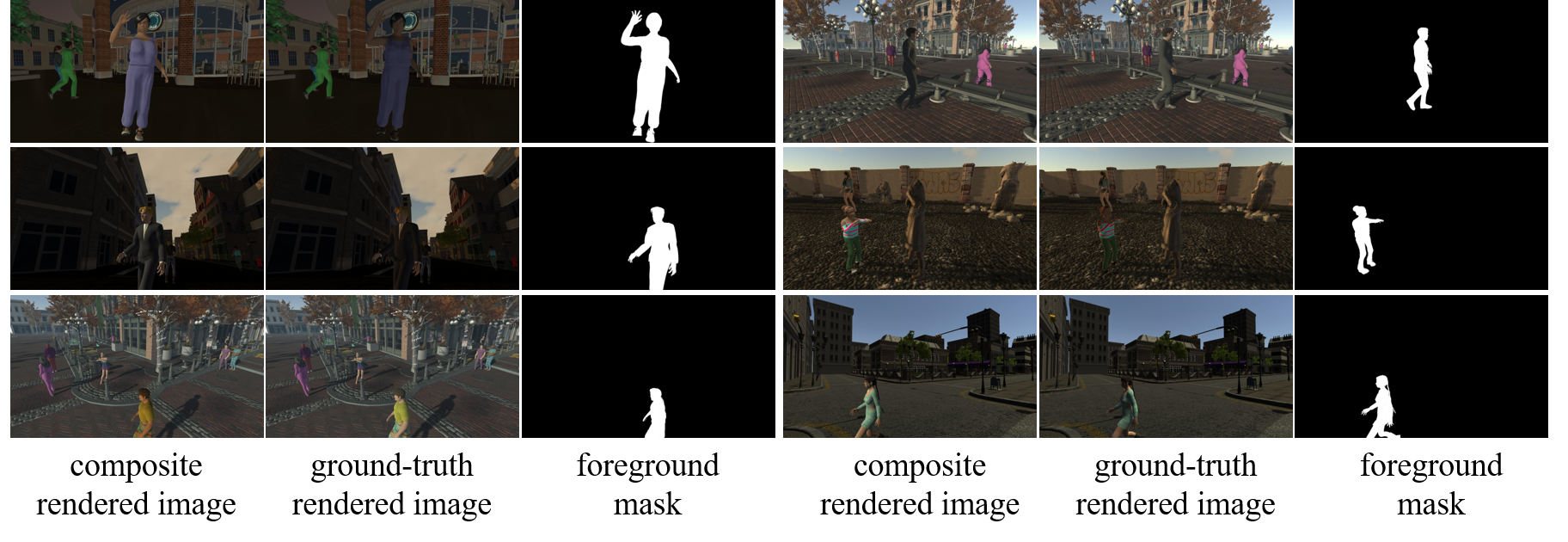}
   \caption{Examples of rendered training pairs from our RdHarmony with ``human'' category.}
\label{fig:image_triplet}
\end{figure*}

\subsection{Composite Rendered Image Generation}\label{sec:compgen}

As shown in Figure~\ref{fig:dataset}, for each 2D scene, we treat one person as foreground and obtain the foreground mask $\mathbf{M}$ effortlessly using Unity3D. We have 10 rendered images $\{\mathbf{I}_i|_{i=1}^{10}\}$ with different styles $\{\mathbf{y}_i|_{i=1}^{10}\}$ from the same 2D scene, where $\mathbf{y}_i$ is a $10$-dim one-hot style label vector. 
We denote the foreground person in rendered image $\mathbf{I}_i$ as $\mathbf{O}_i$. After randomly selecting $\mathbf{I}_i$ and $\mathbf{I}_j$ from $\{\mathbf{I}_i|_{i=1}^{10}\}$, we could generate two pairs of composite rendered images and ground-truth rendered images $\{\mathbf{I}_{i\rightarrow j}, \mathbf{I}_i\}$and $\{\mathbf{I}_{j\rightarrow i}, \mathbf{I}_j\}$ by exchanging $\mathbf{O}_i$ and $\mathbf{O}_j$, where $\mathbf{I}_{j\rightarrow i}$ denotes the composite rendered image with foreground $\mathbf{O}_i$ and background of $\mathbf{I}_j$.

For rendered image pairs $\{\mathbf{I}_{j\rightarrow i}, \mathbf{I}_j\}$, the style label of ground-truth rendered image $\mathbf{I}_j$ is $\mathbf{y}_j$ as mentioned above. For composite rendered image $\mathbf{I}_{j\rightarrow i}$, we simply assume that its style label vector is weighted average of foreground style $\mathbf{y}_i$ and background style $\mathbf{y}_j$ similar to \cite{cutmix}: $\mathbf{y}_{j\rightarrow i} = r\cdot\mathbf{y}_i + (1-r)\cdot\mathbf{y}_j$, in which foreground ratio $r$ denotes the area of foreground over the area of whole image.

We have generated 135,000 pairs of composite rendered images and ground-truth rendered images for ``human'' category in RdHarmony. Examples of rendered training pairs with ``human'' category are shown in Figure \ref{fig:image_triplet}.
It is worth mentioning that each above-mentioned step (\emph{i.e.}, creating 3D characters, capturing 2D scenes, varying capture conditions, and generating composite rendered images) could be done automatically using scripts, making the dataset efficiently constructed and easily extendable. Besides, with controllable capture conditions (\emph{i.e.}, styles) in UniStorm, it is feasible to acquire ground-truth style labels, which are useful for cross-domain knowledge transfer (Section 4.2 in the main text).

\begin{figure*}[htp]
\centering
\includegraphics[width=0.95\linewidth]{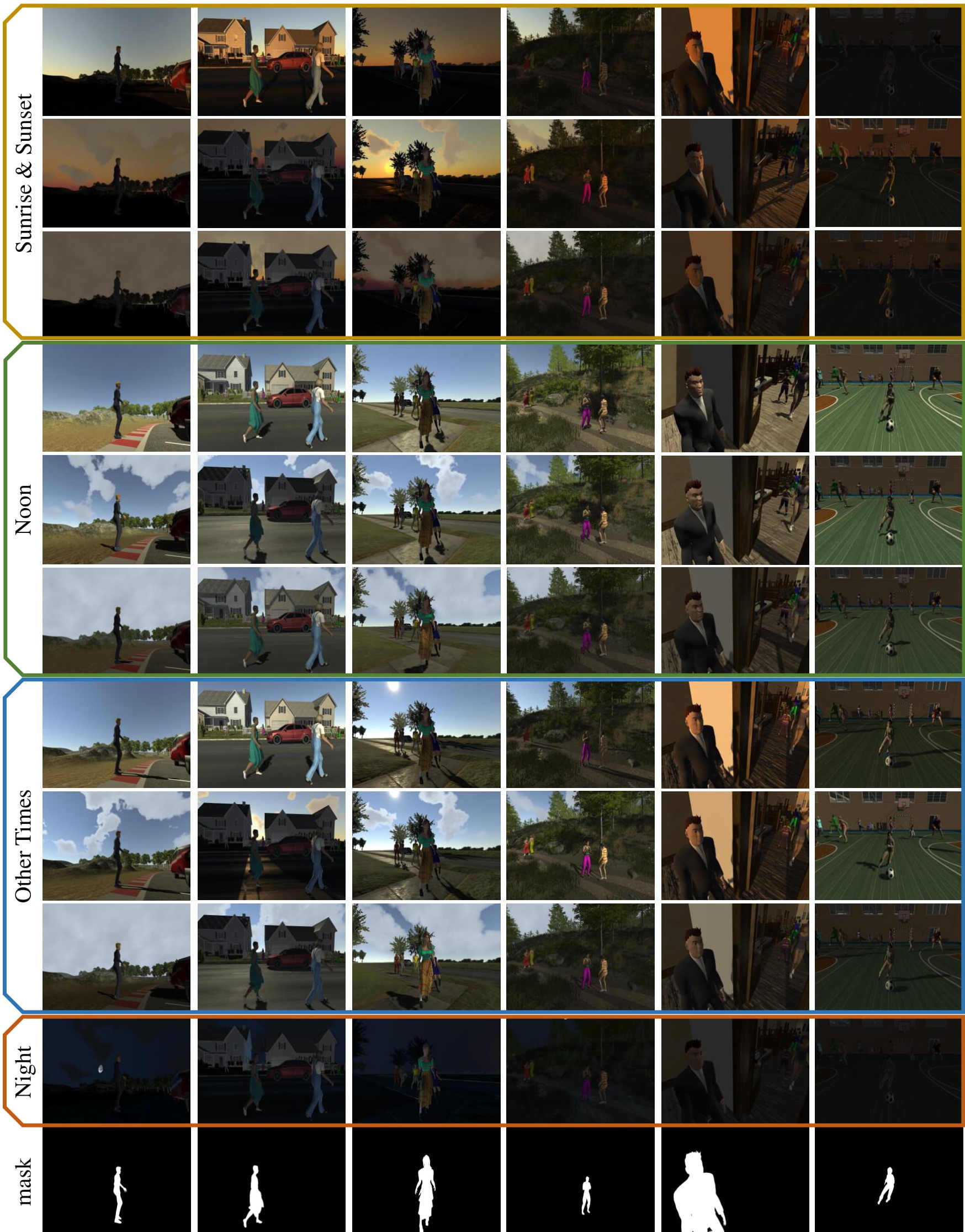}
   \caption{Example ground-truth rendered images in RdHarmony of ``human'' category. The left four columns are outdoor scenes (raceway, downtown, street, and forest) and the right two columns are indoor scenes (bar and stadium). Under each time of the day except ``Night", from top to bottom, we show rendered images captured under Clear, Partly Cloudy, and Cloudy weather.}
\label{fig:rd_example}
\end{figure*}

\begin{figure*}[htp]
\centering
\includegraphics[width=0.95\linewidth]{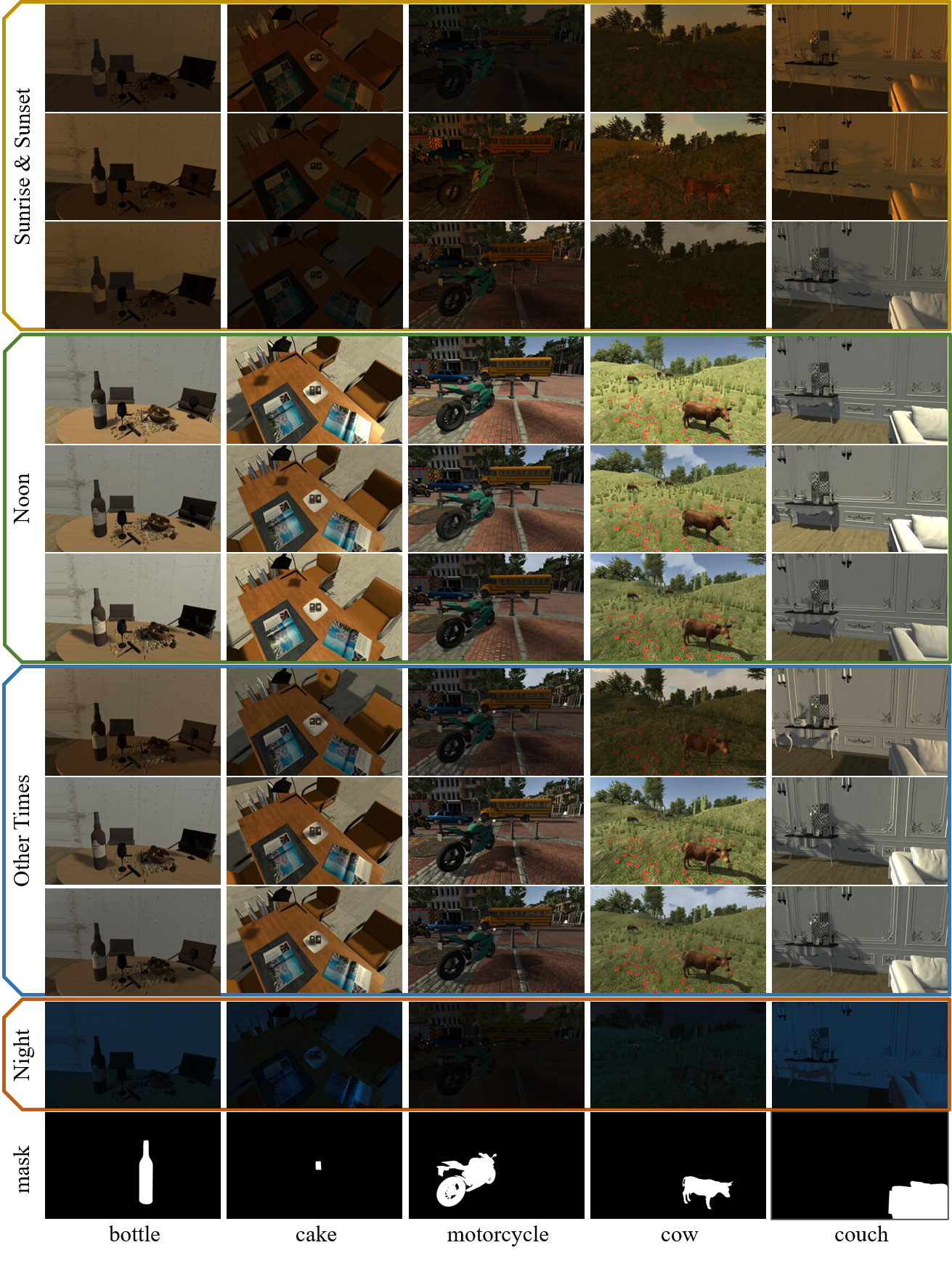}
   \caption{Example ground-truth rendered images of extended 5 object categories. From left to right, we show images from ``bottle'', ``cake'', ``motorcycle'', ``cow'', ``couch'' category. Under each time of the day except ``Night", we show rendered images captured under three representative weather.}
\label{fig:r10_rd_example}
\end{figure*}

\subsection{Considerable Extendability}\label{sec:consider_ext}

We emphasize that we focus on human harmonization in this work, but the dataset construction could be easily generalized to other categories. 
\emph{With the prosperity of Computer Graphics and photogrammetry, 3D model markets are common, making 3D models and 3D scenes more accessible than ever.} To be more specific, 3D models of certain categories could be automatically generated by specific software (\emph{e.g.}, MakeHuman, SpeedTree). 
3D models of other categories can be freely downloaded or purchased from Unity Asset Store and many public websites \footnote{www.cgtrader.com; www.aigei.com; www.cgmodel.com;
www.turbosquid.com; www.sketchfab.com}. 

With 3D scenes and 3D models available, the rest of steps (\emph{i.e.}, capturing 2D scenes, varying capture conditions, and generating composite rendered images) could be done automatically using scripts. To facilitate further study on image harmonization, all scrips used in our dataset construction will be made publicly available. 

Furthermore, for better demonstration, we extend RdHarmony to 5 more categories from 5 super-categories, including ``bottle'', ``cake'', ``motorcycle'', ``cow'', and ``couch''. For each novel category, we place 3D models in each 3D scene, select one 3D model as the foreground and shoot a group of 2D scenes. In detail, for each category, we select 2 3D scenes, and set 50 camera viewpoints for each 3D scene, leading to 100 2D scenes. Then for each 2D scene, we obtain a group of 10 images with 10 different styles. For each category, we can generate 1,000 rendered images with different styles, and produce 9,000 pairs of composite rendered images and ground-truth rendered images. Totally, our RdHarmony has 180,000 training pairs from 6 novel categories.


\subsection{Rich Diversity} \label{sec:human_diversity}
To ensure the diversity of RdHarmony, we have put efforts on the rendered images in several ways. 
First, as mentioned in Section~\ref{sec:compgen}, we collect 30 3D scenes from Unity Asset Store and CG websites, including outdoor scenes (\emph{e.g.}, raceway, downtown, street, forest) and indoor scenes (\emph{e.g.}, bar, stadium, gym). Second, to keep in line with the large intra-category variance of ``human'' category, we leverage MakeHuman to create 1,500 distinct 3D human characters with controllable attributes. For 5 object categories (``bottle'', ``cake'', ``motorcycle'', ``cow'', ``couch''), we collect 3D models from Unity Asset Store and public websites. 
Third, for each 2D scene, we sample 10 ground-truth rendered images with 10 different capture conditions (\emph{i.e.}, styles), including the the night style as well as styles of Clear/Partly Cloudy/Cloudy weather at sunrise\&sunset/noon/other-times. \emph{Note that each style is not a discrete style, but covers a range of illumination intensity and direction.} In Figure~\ref{fig:rd_example}, we select some indoor/outdoor 2D scenes, and exhibit all 10 ground-truth rendered images on ``human'' category for each 2D scene. Besides, Figure \ref{fig:r10_rd_example} shows 10 ground-truth rendered images for each 2D scene and one 2D scene for each object category. These can provide an intuitive perspective for the diversity of our contributed rendered image dataset RdHarmony. 


With a group of 10 images from the same scene, our ground-truth rendered images are quite similar to day2night~\cite{zhou2016evaluating}, \emph{i.e.}, the same scene captured under different capture conditions. Therefore, we naturally adopt the same procedure as constructing Hday2night~\cite{DoveNet2020} when generating composite rendered images, that being said, exchanging foregrounds within the group of images. As stated in \cite{DoveNet2020,guo2021intrinsic}, Hday2night could be deemed as a real composite dataset, where the composite images are much closer to real-world applications. However, collecting a set of images for exactly the same scene under different capture conditions
is not trivial~\cite{DoveNet2020}, which limits the scale of Hday2night. On the contrary, \emph{our dataset construction simulates the ideal generation process of training pairs for image harmonization that could rarely be implemented in real scenarios}. Our RdHarmony, which contains 180,000 pairs, could supplement real dataset to a large extent.


\section{Implementation Details}\label{sec:implement}

We adopt the UNet-like architecture of iDIH as backbone considering its simplicity and effectiveness. Note that we do not use auxiliary semantic information as in~\cite{sofiiuk2021foreground} for brevity and fair comparison with other methods. 
The domain-specific encoder (\emph{resp.}, decoder) contains the first (\emph{resp.}, last) 4 layers in the encoder (\emph{resp.}, decoder). The other layers form the domain-invariant encoder-decoder. 
After a few trials, we set $\lambda_{adv}$, $\lambda_{sty}^{rd}$, and $\lambda_{sty}^{rl}$ to 0.1, 0.1, and 0.05 respectively, by observing the harmonization quality of training images.
The impact of layer configuration and hyper-parameters can be found in Section~\ref{sec:ablation}.
Following~\cite{DoveNet2020}, we use MSE, fMSE (foreground MSE), and PSNR as evaluation metrics. The images are resized to $256 \times 256$ in training and test phases. All the experiments except in Section~\ref{sec:gene_novel} treat ``human'' as the novel category.

Our network is implemented using Pytorch 1.4.0 and trained on ubuntu 16.04 LTS operation system, with 64GB memory, Intel Core i7-8700K CPU, and two GeForce GTX 1080 Ti GPUs. The weight of the network is initialized with values drawn from the normal distribution $\mathcal{N}({mean}=0.0, {std}^2=0.02)$. 

The discriminator $D$ consists of 3 convolution layers with channel number $\{64,32,1\}$, $3\times 3$ kernels and stride of 1, each of which is followed by a Leaky-ReLU except the last one. The style classifiers $P^{in}$ and $P^{out}$ are the same in the structure and shared across domains. They contain 4 convolution layers with channel number $\{64,32,16,1\}$, $3\times 3$ kernels and stride of 1, each of which is followed by a BatchNorm, a ReLU, and a max-pooling layer except the last one.

To train the network, we use Adam optimizer with $\beta_1=0.9$ and $\beta_2=0.999$. Learning rate is initialized with $1e^{-4}$ and reduced by a factor of 10 at epochs 100 and 120. The batch size is set to 8 and the models are trained for 150 epochs.

\section{Ablation Studies}\label{sec:ablation}

In this section, we will validate the effectiveness of different loss terms in Section \ref{sec:loss_design}, explore the variants of network design in Section \ref{sec:network_design}, and analyse the impact of hyper-parameters in Section \ref{sec:aba_hyper}.

\begin{table}[t]
\scriptsize
\begin{center}
\resizebox{\linewidth}{!}{
\begin{tabular}{c|ccccc|cc}
\toprule
 \# & $\mathcal{L}^{rd}_{G}$& $\mathcal{L}^{rd}_{in}$ & $\mathcal{L}^{rd}_{out}$ & $\mathcal{L}^{rl}_{W}$ & $\mathcal{L}^{rl}_{ER}$ & fMSE $\downarrow$ & PSNR $\uparrow$ \\
\hline
1 & & & & & & 339.02 &36.06  \\
2 & \checkmark& & & & & 303.69 &36.45  \\
3 & \checkmark &\checkmark&\checkmark & & & 300.22 & 36.50\\
4 & \checkmark&\checkmark&\checkmark & \checkmark & &  297.99 & 36.52  \\
5 & \checkmark& \checkmark&\checkmark & &\checkmark &  297.94 & 36.51  \\
6 & \checkmark & \checkmark &\checkmark &\checkmark& \checkmark & 296.40 &36.60 \\
\bottomrule
\end{tabular}

\hspace{1em}
\scriptsize
\begin{tabular}{c|c|cc}
\toprule $L$ of $E^{r}$ & $L$ of $D^{r}$ & fMSE$\downarrow$ & PSNR$\uparrow$ \\
\hline
0 & 7 & 498.82 & 34.48 \\
3 & 3 & 327.45 & 36.20 \\
4 & 4 & 296.40 & 36.60\\
5 & 5 & 328.39 & 36.19 \\
7 & 0 & 501.67 & 34.58 \\
\bottomrule
\end{tabular}
\hspace{1em}
}
\end{center}
\caption{Ablation studies. Left sub-table: ablation studies on the losses. Right sub-table: ablation studies on the hyper-parameter $L$. We use $E^{r}$ (\emph{resp.}, $D^{r}$) to represent both $E^{rd}$ (\emph{resp.}, $D^{rd}$) and $E^{rl}$ (\emph{resp.}, $D^{rl}$). ``$L$ of $E^{r}$ (\emph{resp.}, $D^{r}$)'' denotes the number of unshared layers in the first (\emph{resp.}, third) stage. All results are tested on $\mathcal{D}^{rl}_{te,n}$.}
\label{tab:ablate_gen}
\end{table}

\subsection{Loss Design} \label{sec:loss_design}
We ablate each component of the loss function in Table \ref{tab:ablate_gen}. When we use neither adversarial loss nor style-related losses, the network only contains two three-stage generators. Though the performance is degraded, it is still better than iDIH (row 4 in Table 2 in the main text) due to the information sharing in the second stage. By adding adversarial loss $\mathcal{L}^{rd}_{G}$ after the first stage, the performance is boosted, which demonstrates the efficacy of adversarial loss to pull close two domains. Moreover, after adding two style losses $\mathcal{L}^{rd}_{in}$ and $\mathcal{L}^{rd}_{out}$ in the rendered image domain, the performance is further improved, which indicates the potential of available style information. We ablate the style aggregation loss $\mathcal{L}^{rl}_{SA}$ by its two terms $\mathcal{L}^{rl}_{W}$ and $\mathcal{L}^{rl}_{ER}$. The results demonstrate that each term is helpful and their combination achieves further improvement with mutual collaboration.

\begin{figure*}[t]
\centering
\includegraphics[width=0.98\linewidth]{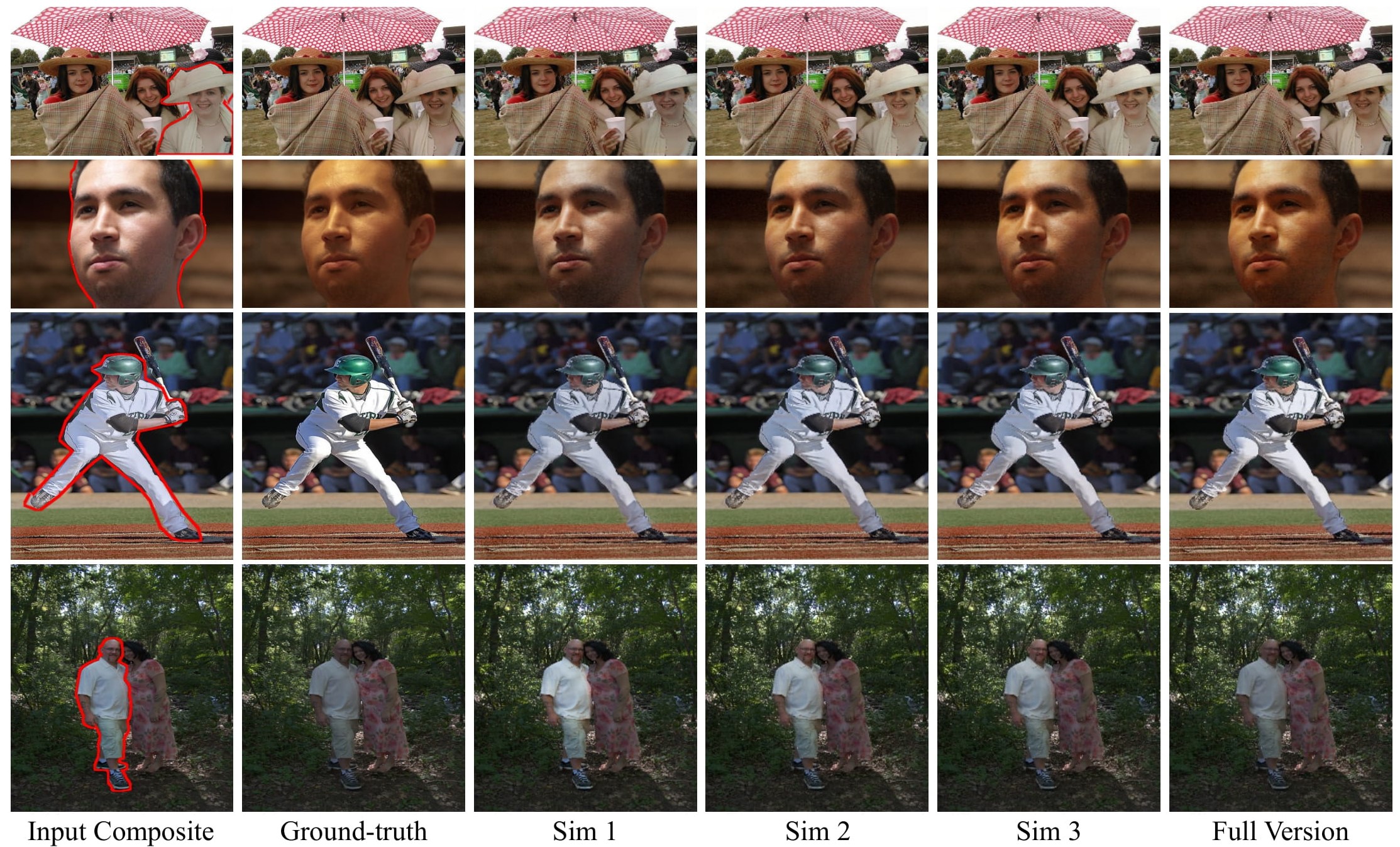}
   \caption{Example results generated under different loss designs on $\mathcal{D}^{rl}_{te,n}$. From left to right, we show the input composite real image, ground-truth real image, as well as the harmonized results generated by Sim 1, Sim 2, Sim 3, and our full method CharmNet, where Sim 1 $=\mathcal{L}^{rd}_{rec}+\mathcal{L}^{rl}_{rec}$, Sim 2 $=\mathcal{L}^{rd}_{rec}+\mathcal{L}^{rl}_{rec}+\mathcal{L}^{rd}_{G}$, Sim 3 $= \mathcal{L}^{rd}_{rec}+\mathcal{L}^{rl}_{rec}+\mathcal{L}^{rd}_{G} +\mathcal{L}^{rd}_{in}+\mathcal{L}^{rd}_{out}$, and Full Version $=\mathcal{L}^{rd}_{rec}+\mathcal{L}^{rl}_{rec}+\mathcal{L}^{rd}_{G} +\mathcal{L}^{rd}_{in}+\mathcal{L}^{rd}_{out}+\mathcal{L}^{rl}_{SA}$. The foregrounds are outlined in red.}
\label{fig:loss_visual}
\end{figure*}

We present example test images from $\mathcal{D}^{rl}_{te,n}$ harmonized under four different loss designs, including $\mathcal{L}^{rd}_{rec}+\mathcal{L}^{rl}_{rec}$ (row 1 of left sub-table in Table \ref{tab:ablate_gen}), $\mathcal{L}^{rd}_{rec}+\mathcal{L}^{rl}_{rec}+\mathcal{L}^{rd}_{G}$ (row 2), $\mathcal{L}^{rd}_{rec}+\mathcal{L}^{rl}_{rec}+\mathcal{L}^{rd}_{G}+\mathcal{L}^{rd}_{in}+\mathcal{L}^{rd}_{out}$ (row 3), and our full method $\mathcal{L}^{rd}_{rec}+\mathcal{L}^{rl}_{rec}+\mathcal{L}^{rd}_{G}+\mathcal{L}^{rd}_{in}+\mathcal{L}^{rd}_{out}+\mathcal{L}^{rl}_{SA}$ (row 6). The results are shown in Figure~\ref{fig:loss_visual}. We observe that our CharmNet could produce more harmonious results that are closer to the ground-truth real images. By comparing the results of ``Sim 1" and ``Sim 2" in Figure~\ref{fig:loss_visual}, it can be seen that adversarial loss $\mathcal{L}^{rd}_{G}$ is important for aligning two domains to generate reasonable harmonized results. By comparing the results of ``Sim 2", ``Sim 3" and ``Full Version", it can be seen that style classification losses $\mathcal{L}^{rd}_{in}+\mathcal{L}^{rd}_{out}$ in the rendered image domain and the style aggregation loss $\mathcal{L}^{rl}_{SA}$ in the real image domain are both meaningful for image harmonization, and their combination ensures the useful knowledge transfer across domains, thus leading to more satisfactory performance.

\subsection{Network Design} \label{sec:network_design}
We analyze the network design in terms of the number of unshared layers in the domain-specific encoding stage and domain-specific decoding stage. 
As mentioned in Section 4.1 in the main text, we empirically split the first (\emph{resp.}, last) $L$ layers in encoder (\emph{resp.}, decoder) into the first (\emph{resp.}, third) stage. Therefore, we also investigate the impact of hyper-parameter $L$, and the results are also reported in Table \ref{tab:ablate_gen}. When we increase $L$ to 5, the performance is degraded, since it hampers the useful information sharing in the second stage. In addition, when we decrease $L$ to 3, the performance is also degraded, possibly because it brings difficulty in aligning the extracted features from two domains. 
We also explore the extreme cases where all the encoder or decoder layers are shared across domains. Setting $L$ of $E^{r}$ (\emph{resp.}, $D^{r}$) to 0 means sharing the entire encoder (\emph{resp.}, decoder). The performances in both cases are significantly dropped. This is because the input space and output space for image harmonization are the same in each domain but different across domains. Simply sharing the entire encoder or decoder will hurt the cross-domain transfer.

\begin{figure}[htb]
\centering
\includegraphics[width=.65\linewidth]{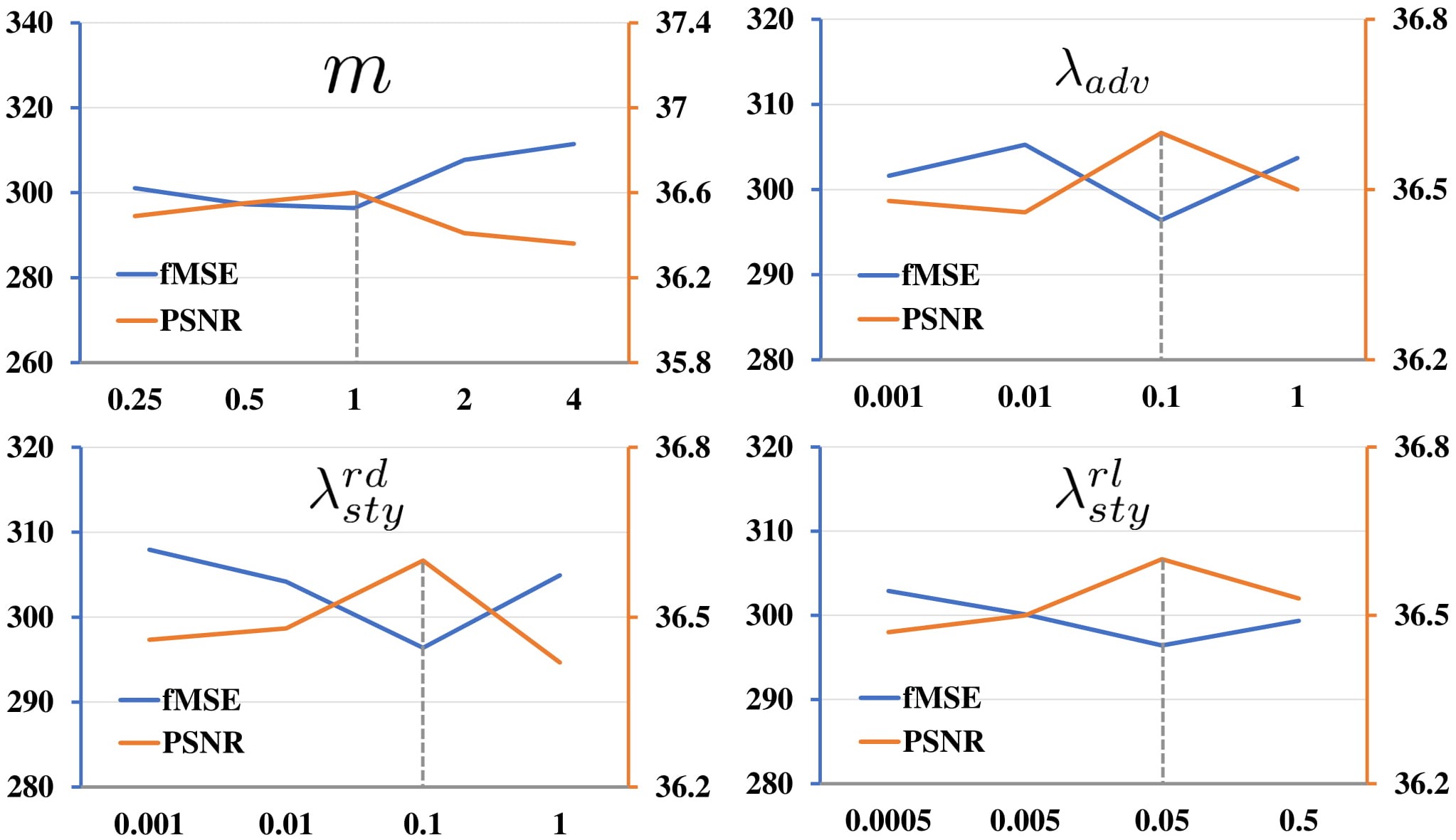}
\caption{Impact of hyper-parameters, including the margin $m$ in Eqn. 5 and trade-off parameters $\lambda_{adv}$, $\lambda_{sty}^{rd}$, and $\lambda_{sty}^{rl}$ in Eqn. 6 in the main text. The results are tested on $\mathcal{D}^{rl}_{te,n}$ and the gray dotted line indicates the default value of each hyper-parameter.}
\label{fig:hyperparam}
\end{figure}


\subsection{Hyper-parameter Analyses}\label{sec:aba_hyper}

We investigate the impact of four hyper-parameters: the margin $m$ in Eqn. 5 and trade-off parameters $\lambda_{adv}$, $\lambda_{sty}^{rd}$, and $\lambda_{sty}^{rl}$ in Eqn. 6 in the main paper. In Figure~\ref{fig:hyperparam}, we plot the performance on $\mathcal{D}^{rl}_{te,n}$ by varying each hyper-parameter while keeping the other hyper-parameters fixed. It can be seen that our method is robust with hyper-parameters in reasonable ranges (\emph{i.e.}, $m$ in range $[2^{-2},2^{2}]$, $\lambda _{adv}$ and $\lambda_{sty}^{rd}$ in range $[10^{-3},1]$ and $\lambda_{sty}^{rl}$ in range $[5\times 10^{-4},5\times 10^{-1}]$). 

\begin{figure}[t]
\centering
\subfigure[Train on different numbers of scenes. Test on $\mathcal{D}^{rl}_{te,n}$.]{
\label{fig:scene}
\includegraphics[width=.37\linewidth]{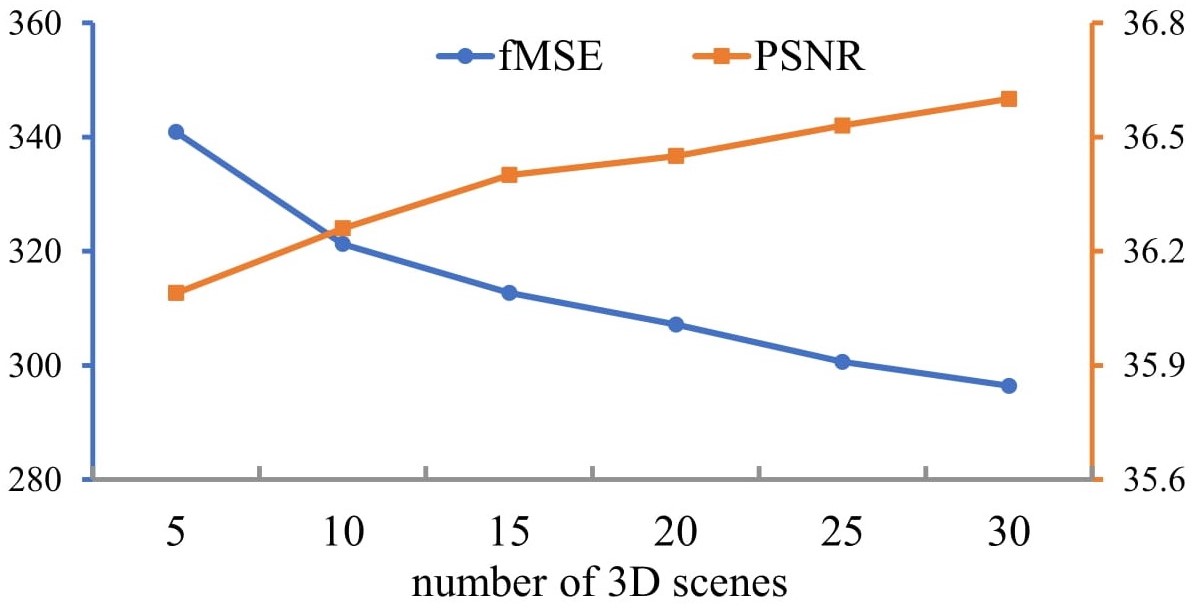}
}
\quad
\subfigure[Train on different numbers of styles. Test on $\mathcal{D}^{rl}_{te,n}$.]{
\label{fig:style}
\includegraphics[width=.37\linewidth]{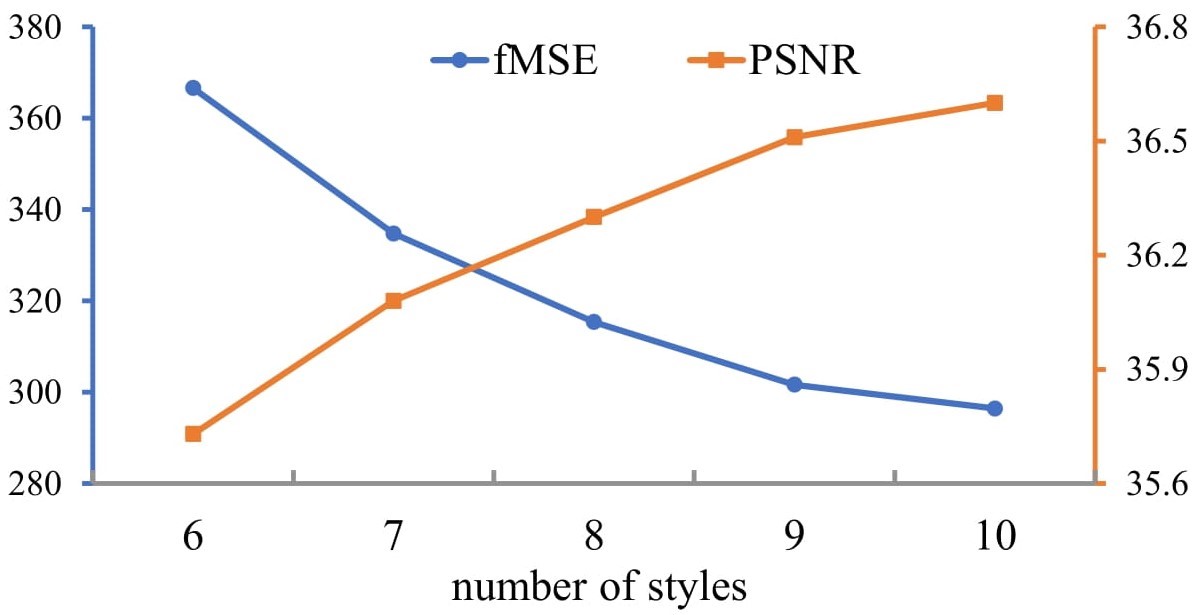}
}
\quad
\subfigure[Train on $\mathcal{D}^{rl}_{tr,b}+\mathcal{D}^{rd}_{tr,n}(sub)$. Test on $\mathcal{D}^{rl}_{te,n}$.]{
\label{fig:render_novel}
\includegraphics[width=.37\linewidth]{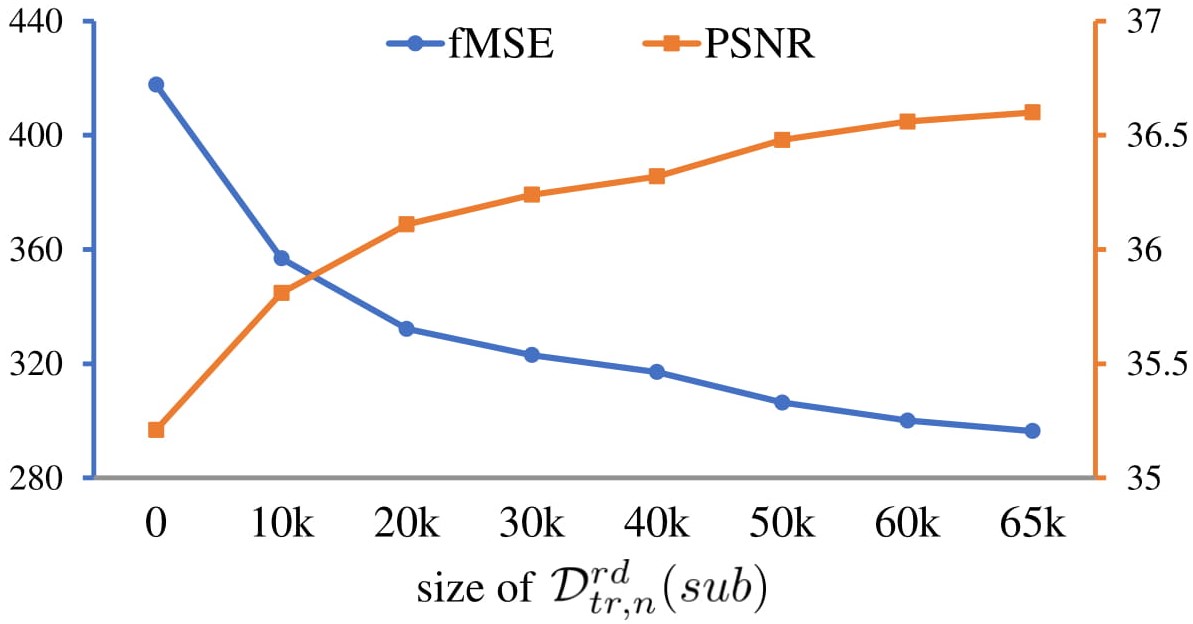}
}
\quad
\subfigure[Train on $\mathcal{D}^{rl}_{tr,b}+\mathcal{D}^{rd}_{tr,n}(sub)$. Test on $\mathcal{D}^{rl}_{te,b}$.]{
\label{fig:render_base}
\includegraphics[width=.37\linewidth]{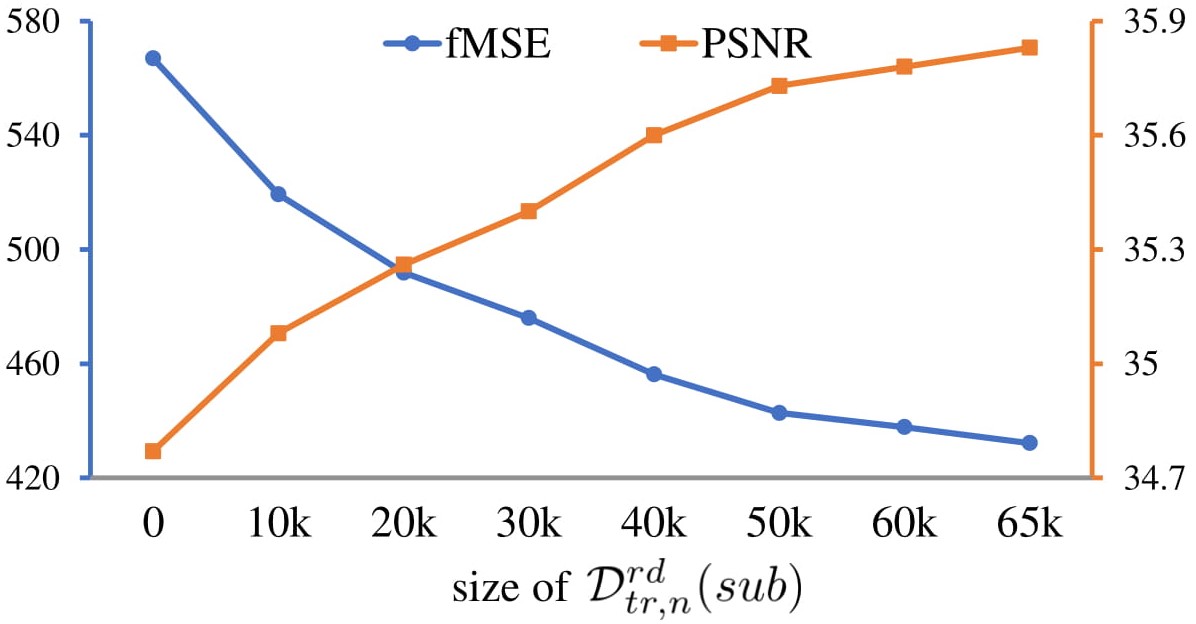}
}
\quad
\subfigure[Train on $\mathcal{D}^{rl}_{tr,b}+\mathcal{D}^{rl}_{tr,n}(sub)$ with and without $\mathcal{D}^{rd}_{tr,n}$. Test on $\mathcal{D}^{rl}_{te,n}$.]{
\label{fig:real_novel}
\includegraphics[width=.37\linewidth]{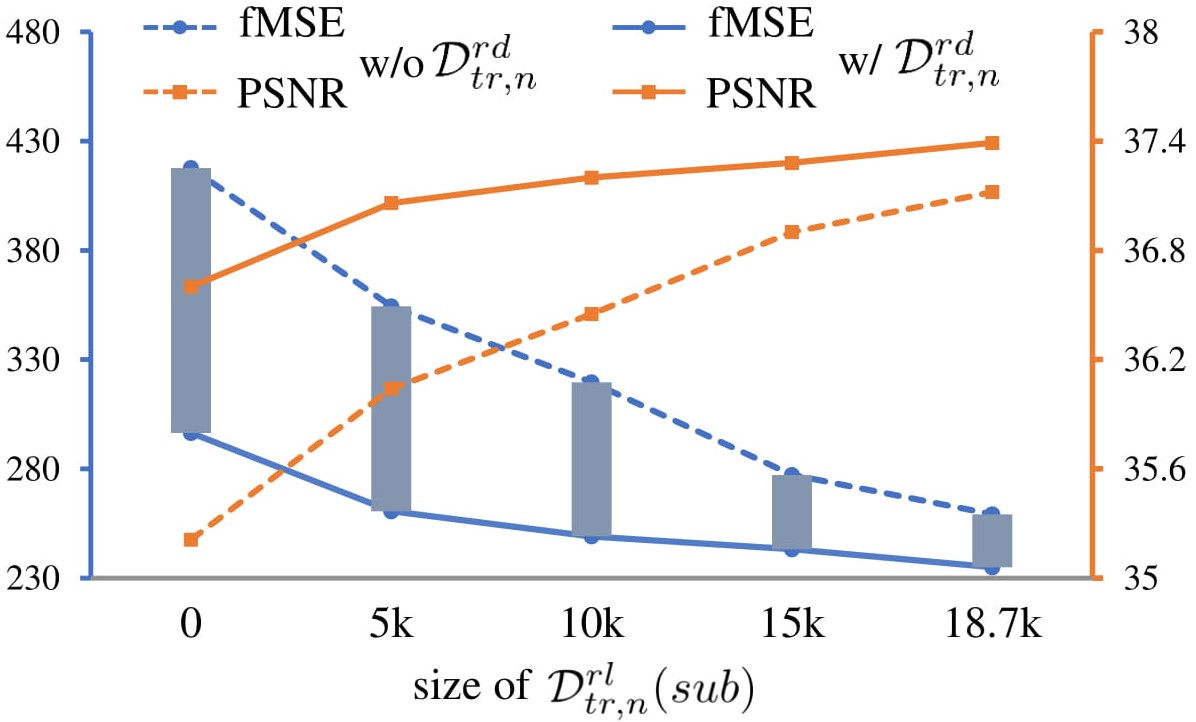}
}
\quad
\subfigure[Train on $\mathcal{D}^{rl}_{tr,b}+\mathcal{D}^{rl}_{tr,n}(sub)$ with and without $\mathcal{D}^{rd}_{tr,n}$. Test on $\mathcal{D}^{rl}_{te,b}$.]{
\label{fig:real_base}
\includegraphics[width=.37\linewidth]{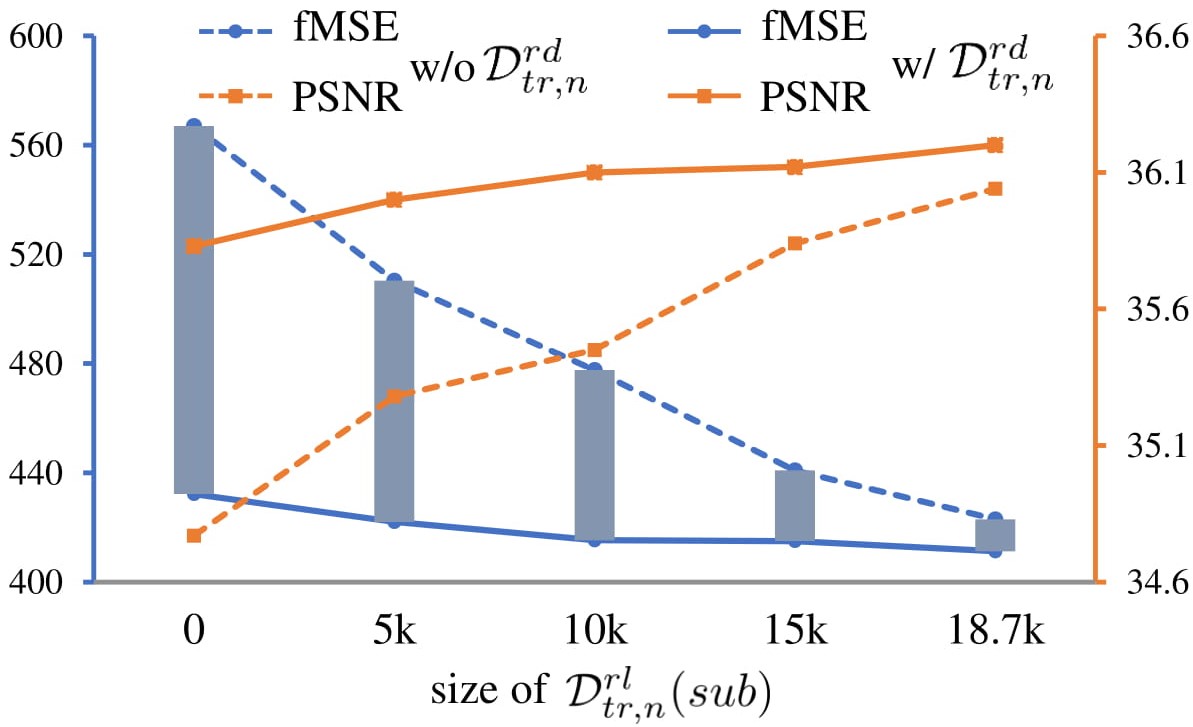}
}
\caption{
Top row: training with a subset of 10 styles and 30 3D scenes. 
Mid row: training with a subset of $\mathcal{D}^{rd}_{tr,n}$ (\emph{i.e.}, $\mathcal{D}^{rd}_{tr,n}(sub)$) and the full set of $\mathcal{D}^{rl}_{tr,b}$. 
Bottom row: training with a subset of $\mathcal{D}^{rl}_{tr,n}$ (\emph{i.e.}, $\mathcal{D}^{rl}_{tr,n}(sub)$) and the full set of $\mathcal{D}^{rl}_{tr,b}$ with and without the full set of $\mathcal{D}^{rd}_{tr,n}$. 
}
\label{fig:traindata}
\end{figure}

\section{Training Data Analyses}\label{sec:supp_traindata}

In the dataset construction, to ensure the diversity, we collect 30 3D scenes to cover various virtual environments and define 10 representative styles to cover the majority of the day (we do not sample the whole night since there is little variation when it is totally dark) and 3 representative weather. Note that each style covers a range of illumination intensity and direction. 
To investigate the influence of diversity of constructed RdHarmony, we explore the performance variance by using different numbers of 3D scenes and styles. To be specific, for 3D scenes, we use the full set of $\mathcal{D}^{rl}_{tr,b}$ and a subset of $\mathcal{D}^{rd}_{tr,n}$ (\emph{i.e.}, $\mathcal{D}^{rd}_{tr,n}(sub)$) which belongs to a subset of 30 scenes. Similarly, for styles, we use $\mathcal{D}^{rl}_{tr,b}$ and $\mathcal{D}^{rd}_{tr,n}(sub)$ which contains a subset of 10 styles. The results are shown in Figure \ref{fig:scene} and Figure \ref{fig:style}. From Figure \ref{fig:scene}, we can observe that the performance grows better when using more scenes. As the number of 3D scenes increases up to 30,  we can also observe a trend in convergence. Since for each 3D scene we set 50 camera viewpoints to shot 2D scenes, an increase in 3D scene numbers leads to a much larger increase in 2D scene numbers, which makes the dataset more and more diverse. From Figure \ref{fig:style}, we have a similar observation. When using fewer styles, since it could not cover both time-of-the-day and weathers well, the performance is inferior. When the number of styles increases to 10, the performance is also improved and getting stable at a superior level. Based on such observation, we use all 30 3D scenes and 10 styles in all experiments which use $\mathcal{D}^{rd}_{tr,n}$.

In the experiments in the main paper, we use $65k$ rendered training images with ``human'' category in $\mathcal{D}^{rd}_{tr,n}$. Here, we explore the performance variance when using different numbers of rendered training images. Specifically, we use the full set of $\mathcal{D}^{rl}_{tr,b}$ and a subset of $\mathcal{D}^{rd}_{tr,n}$ (\emph{i.e.}, $\mathcal{D}^{rd}_{tr,n}(sub)$) with various numbers of rendered training images to train our CharmNet, and evaluate the model on both $\mathcal{D}^{rl}_{te,n}$ and $\mathcal{D}^{rl}_{te,b}$. As shown in Figure~\ref{fig:render_novel} and Figure~\ref{fig:render_base}, the performance is improved with an increasing size of $\mathcal{D}^{rd}_{tr,n}(sub)$. Though the performance growth slows down when the size approaches $65k$, the best performance is still obtained under the full set of $\mathcal{D}^{rd}_{tr,n}$. Note that introducing more rendered training images will bring marginal improvement yet higher computational cost, so we use $65k$ rendered training images by default.

Besides, we explore another experimental setup, where real training set has inadequate number of examples from novel category. \emph{It is very common that the real training set has a few yet insufficient examples for certain categories, in which case we can enrich the real training set with rendered images from these categories.} Specifically, we use the full set of $\mathcal{D}^{rl}_{tr,b}$ and a subset of $\mathcal{D}^{rl}_{tr,n}$ (\emph{i.e.}, $\mathcal{D}^{rl}_{tr,n}(sub)$) with various numbers of real training images from novel category to train the iDIH~\cite{sofiiuk2021foreground} backbone. In addition, for each size of $\mathcal{D}^{rl}_{tr,n}(sub)$, we also augment the training set with the full set of $\mathcal{D}^{rd}_{tr,n}$ to train our CharmNet. The trained models are evaluated on both $\mathcal{D}^{rl}_{te,n}$ and $\mathcal{D}^{rl}_{te,b}$ for comparison. The results are shown in Figure~\ref{fig:real_novel} and Figure~\ref{fig:real_base}, from which we have three observations. Firstly, no matter using rendered images or not, increasing the size of $\mathcal{D}^{rl}_{tr,n}(sub)$ will bring consistent performance improvement. Secondly, with a specific size of $\mathcal{D}^{rl}_{tr,n}(sub)$ (\emph{e.g.}, $5k$), using rendered training images $\mathcal{D}^{rd}_{tr,n}$ from novel category could boost the performance. Thirdly, when the size of $\mathcal{D}^{rl}_{tr,n}(sub)$ is small, the performance gain brought by $\mathcal{D}^{rd}_{tr,n}$ and our CharmNet is significant, which demonstrates that \emph{auxiliary rendered images are especially helpful when the real training set is short of real images from novel categories.}

\begin{figure*}[t]
\centering
\includegraphics[width=.95\linewidth]{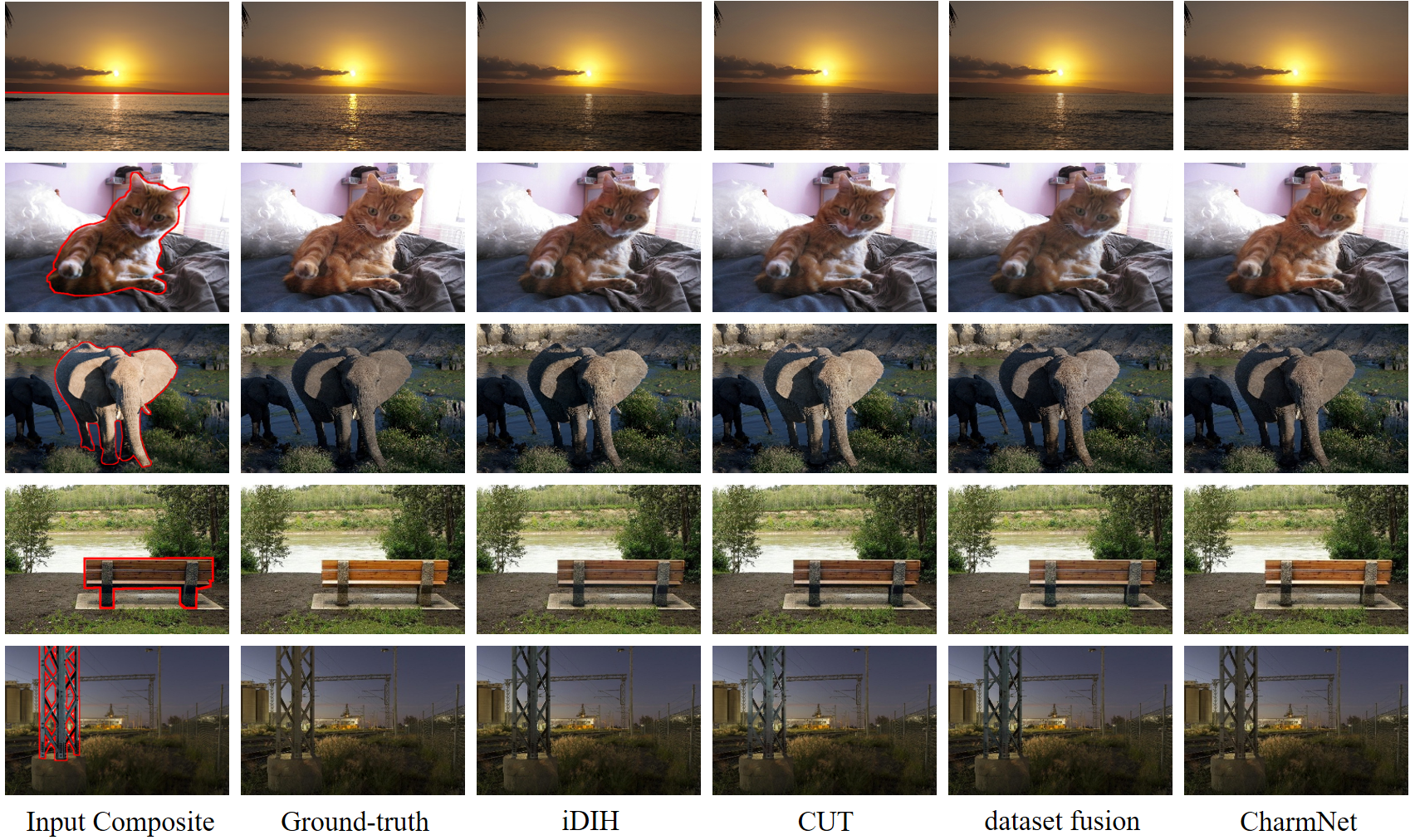}
   \caption{Example results generated by different baselines and our method on $\mathcal{D}^{rl}_{te,b}$. From left to right, we show the input composite real image, ground-truth real image, as well as the harmonized images generated by iDIH~\cite{sofiiuk2021foreground} backbone, CUT~\cite{park2020cut}, dataset fusion, and our CharmNet. The foregrounds are outlined in red.}
\label{fig:examples_base}
\end{figure*}

\begin{figure}[htb]
\centering
\includegraphics[width=.7\linewidth]{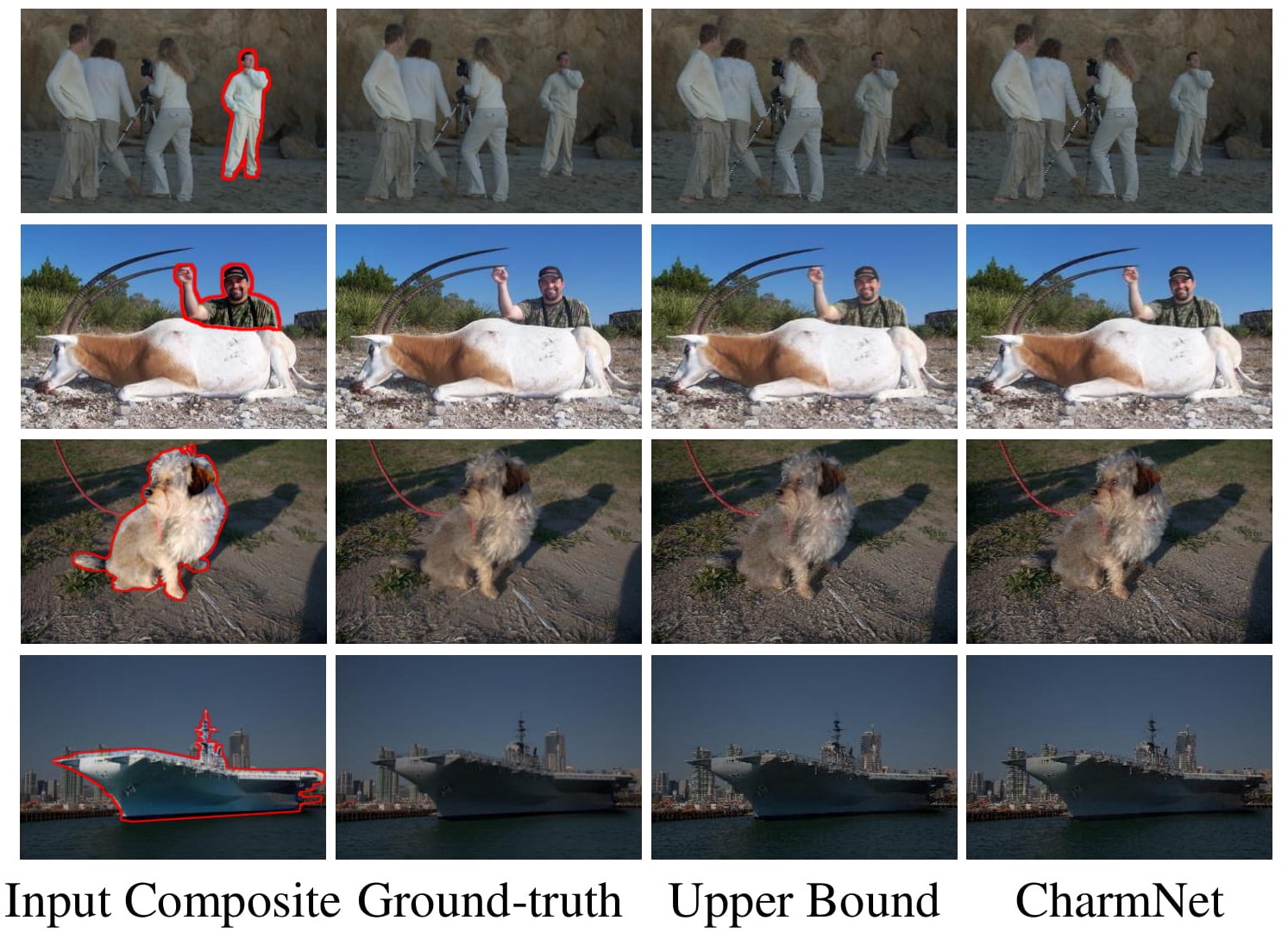}
\caption{Example results generated by iDIH upper bound and our method. The top two rows are images from $\mathcal{D}^{rl}_{te,n}$ and the bottom two rows are images from $\mathcal{D}^{rl}_{te,b}$. From left to right, we show the input composite real image, ground-truth real image, as well as the harmonized images generated by the upper bound and our CharmNet. Foregrounds are outlined in red.}
\label{fig:upperbound}
\end{figure}

\section{Qualitative Analyses on Base Categories}\label{sec:res_real_base}

In Section 5.3 in the main text, we observe that our CharmNet not only boosts the performance on novel category, but also significantly enhances the performance on base categories. Therefore, given a composite real image from $\mathcal{D}^{rl}_{te,b}$, we also provide the harmonization results generated by iDIH~\cite{sofiiuk2021foreground} (row 4 in Table 2 in the main text), CUT~\cite{park2020cut}, dataset fusion, and our CharmNet for comparison. As shown in Figure~\ref{fig:examples_base}, harmonization results of our CharmNet are more plausible and closer to the ground-truth real images. It could be observed that even if the foreground is not ``human", the foreground style could be well-adapted to the background style, which indicates that \emph{the cross-domain style knowledge transfer is also useful for cross-category harmonization.} This observation also demonstrates the potential of using rendered images for image harmonization task.

\section{Comparison between Our CharmNet and the Upper Bound}\label{sec:upper}

In Table 2 in the main text, our CharmNet outperforms all the domain adaptation baselines and achieves a closer performance to the iDIH~\cite{sofiiuk2021foreground} trained with both $\mathcal{D}^{rl}_{tr,n}$ and $\mathcal{D}^{rl}_{tr,b}$, which serves as an upper bound. 
Therefore, we sample real test images from both $\mathcal{D}^{rl}_{te,n}$ and $\mathcal{D}^{rl}_{te,b}$ and show the ground-truth real images as well as the harmonized results of our CharmNet and the upper bound in Figure~\ref{fig:upperbound}. It can be observed that \emph{our CharmNet could generate harmonious results close to the upper bound for both novel and base categories}, which demonstrates the efficacy of our cross-domain harmonization method.

\section{Results on Real-world Composite Images with Human Foregrounds}\label{sec:real_comp}

In practice, image harmonization is expected to tackle with real-world composite images, whose foreground is cut from one image and pasted on another background image. In such a scenario, there are no corresponding ground-truth images, so it is infeasible to evaluate model performance quantitatively. Following \cite{tsai2017deep,xiaodong2019improving,DoveNet2020}, we conduct user study on $46$ real-world composite images with human foregrounds which are selected from $99$ real-world composite images released by \cite{tsai2017deep}, and compare the harmonized results generated by iDIH~\cite{sofiiuk2021foreground} (row 4 in Table 2 in the main text), CUT~\cite{park2020cut}, dataset fusion, and our CharmNet using subjective evaluation.

Specifically, for each real-world composite image, we could obtain five images $\{\mathbf{I}_i|_{i=1}^5\}$ including itself and four harmonized outputs generated by four above-mentioned methods. Then we can construct image pairs $(\mathbf{I}_i, \mathbf{I}_j)$ by randomly selecting two images from $\{\mathbf{I}_i|_{i=1}^5\}$. Based on $46$ real-world composite images with human foregrounds, we could construct abundant image pairs for user study. Then we invite $13$ users to participate in the study. We ask each user to see one image pair each time and pick out the more harmonious image in the pair. Finally, we collect $5980$ pairwise results and employ the Bradley-Terry (B-T) model~\cite{bradley1952rank,lai2016comparative} to obtain the overall ranking of all methods. As reported in Table~\ref{tab:BT_score}, our CharmNet achieves the highest B-T score and once again outperforms other baselines.

\begin{figure*}[htp!]
\centering
\includegraphics[width=0.95\linewidth]{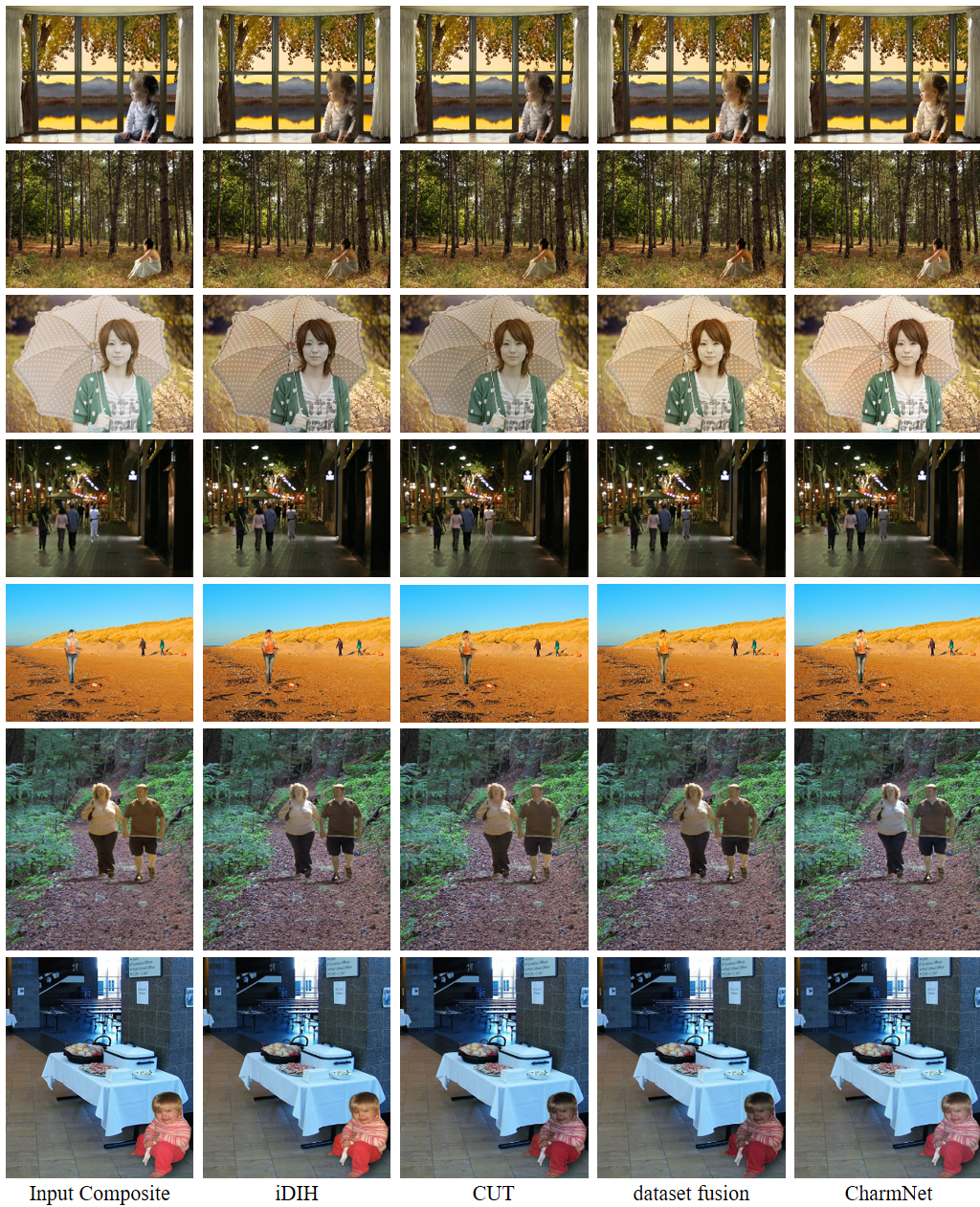}
   \caption{Example results generated by different baselines and our method on real-world composite images with human foregrounds. From left to right, we show the input real-world composite image and the harmonized results generated by iDIH~\cite{sofiiuk2021foreground} backbone (row 4 in Table 2 in the main paper), CUT~\cite{park2020cut}, dataset fusion, and our CharmNet.}
\label{fig:realcomp}
\end{figure*}

Besides, we exhibit some example results of real-world composite images with human foregrounds used in our user study. We compare the real-world composite images with harmonization results generated by iDIH~\cite{sofiiuk2021foreground} (row 4 in Table 2 in the main text), CUT~\cite{park2020cut}, dataset fusion, and our CharmNet. As shown in Figure ~\ref{fig:realcomp}, our method is capable of generating more favorable and satisfactory results than other methods.

\begin{table}[t]
\setlength\tabcolsep{4pt}
\small
\begin{center}
\begin{tabular}{c|ccccc}
\toprule
Method & Input composite & iDIH~\cite{sofiiuk2021foreground} & CUT~\cite{park2020cut} & dataset fusion & CharmNet \\
\hline
B-T score$\uparrow$ & -1.191 & 0.250 & 0.137 & 0.326 & 0.479 \\
\bottomrule
\end{tabular}
\end{center}
\caption{B-T scores of different methods and our CharmNet on $46$ real-world composite images with human foregrounds.}
\label{tab:BT_score}
\end{table}

\begin{table*}[t]
\setlength\tabcolsep{7pt}
\small
\begin{center}
\begin{tabular}{c|c|c|ccc|c}
\toprule
\multicolumn{2}{c|}{\#} & 1 & 2 & 3 & 4 & 5 \\ \hline
\multicolumn{2}{c|}{Training data} & - & $\mathcal{D}^{rl}_{tr,b}$ & \makecell{$\mathcal{D}^{rd}_{tr,n(H)}$ \\ \&  $\mathcal{D}^{rl}_{tr,b}$} & \makecell{$\mathcal{D}^{rd}_{tr,n}$ \\ \&  $\mathcal{D}^{rl}_{tr,b}$} & \makecell{$\mathcal{D}^{rl}_{tr,n}$ \\ \&  $\mathcal{D}^{rl}_{tr,b}$} \\
\hline
\multirow{2}*{$\mathcal{D}^{rl}_{te,n(Cake)}$} & fMSE$\downarrow$ & 1293.83 & 439.33 & 465.49 & 425.63 & 391.74 \\
~ & PSNR$\uparrow$ & 29.72 & 33.93 & 33.71 & 34.07 & 34.43 \\
\hline
\multirow{2}*{$\mathcal{D}^{rl}_{te,n(Bottle)}$} & fMSE$\downarrow$ & 1604.52 & 661.81 & 656.73 & 626.48 & 622.39 \\
~ & PSNR$\uparrow$ & 32.78 & 35.69 & 35.83 & 36.01 & 36.05 \\
\hline
\multirow{2}*{$\mathcal{D}^{rl}_{te,n(Motorcycle)}$} & fMSE$\downarrow$ & 936.30 & 326.49 & 322.99 & 315.62 & 285.76 \\
~ & PSNR$\uparrow$ & 32.28 & 35.91 & 35.94 & 36.07 & 36.56 \\
\hline
\multirow{2}*{$\mathcal{D}^{rl}_{te,n(Cow)}$} & fMSE$\downarrow$ & 921.92 & 411.46 & 401.94 & 374.70 & 382.79 \\
~ & PSNR$\uparrow$ & 34.40 & 37.36 & 37.42 & 37.53 & 37.80 \\
\hline
\multirow{2}*{$\mathcal{D}^{rl}_{te,n(Couch)}$} & fMSE$\downarrow$ & 1565.55 & 575.16 & 563.49 & 531.70 & 504.99 \\
~ & PSNR$\uparrow$ & 29.66 & 33.09 & 33.41 & 33.79 & 33.81 \\
\bottomrule
\end{tabular}
\end{center}
\caption{Results of CharmNet trained on various training data and tested on $\mathcal{D}^{rl}_{te,n}$. ``-" denotes metrics directly tested on composite real images. $\mathcal{D}^{rd}_{tr,n(H)}$ denotes rendered training images from ``human'' category.
Note that column 5 using real training images $\mathcal{D}^{rl}_{tr,n}$ from novel category serves as the upper bound.}

\label{tab:novel_five}
\end{table*}

\begin{figure*}[t]
\centering
\includegraphics[width=0.95\linewidth]{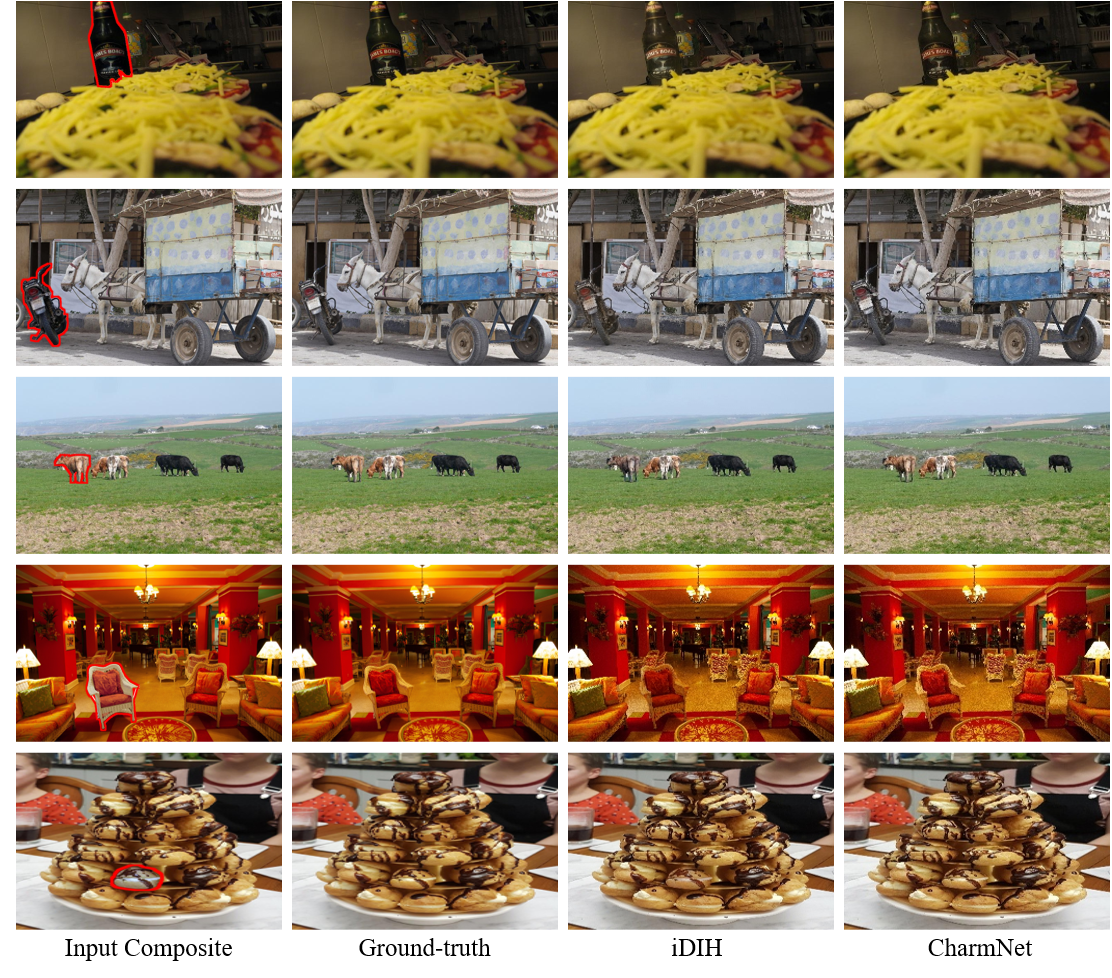}
   \caption{Example results on 5 more novel categories. From top to bottom, we show one example from ``bottle'', ``motorcycle'', ``cow'', ``couch'', and ``cake'' category respectively. From left to right, we show the input composite real image, ground-truth real image, as well as the harmonized images generated by iDIH~\cite{sofiiuk2021foreground} backbone, and our CharmNet. The foregrounds are outlined in red.}
\label{fig:novel_five_res}
\end{figure*}

\section{Experiments on More Novel Categories}\label{sec:gene_novel}
In this paper, we mainly take ``human" as the example novel category to demonstrate the effectiveness of our CharmNet. In this section, we conduct experiments on more novel categories to show the generalization ability of our method.
As mentioned in Section~\ref{sec:consider_ext}, we extend our RdHarmony dataset to 5 more categories (``bottle'', ``cake'', ``motorcycle'', ``cow'', and ``couch'') from 5 different super-categories. Each category has 9,000 pairs of composite rendered images and ground-truth rendered images. Next, we conduct experiments by using these 5 categories as example novel categories one by one and the procedure is basically the same as that with ``human" novel category. 
By taking ``bottle" as an example novel category, we treat the other categories as base categories. The training set ($\mathcal{D}^{rd}_{tr,n}$ \& $\mathcal{D}^{rl}_{tr,b}$) and the test sets ($\mathcal{D}^{rl}_{te,n}$ and $\mathcal{D}^{rl}_{te,b}$) of ``bottle'' novel category are obtained following the same procedure as those of ``human'' novel category in the main text. Our training set consists of the novel training set  $\mathcal{D}^{rd}_{tr,n}$ with rendered images from ``bottle" category and the base training set  $\mathcal{D}^{rl}_{tr,b}$ with real training images from other categories. We evaluate on the novel test set $\mathcal{D}^{rl}_{te,n}$ with real test images from ``bottle" category. The results of this setting are reported in column 4 in Table \ref{tab:novel_five}.
We compare with the results only using base training set  $\mathcal{D}^{rl}_{tr,b}$ (column 2). To prove that within-category cross-domain transfer is more effective than cross-category cross-domain transfer, we also replace the novel training set  $\mathcal{D}^{rd}_{tr,n}$ with equal number of rendered images from ``human" category (9,000 images from RdHarmony) and report the results in column 3. Finally, we report the upper bound results using both base training set  $\mathcal{D}^{rl}_{tr,b}$ and real training images from ``bottle" category $\mathcal{D}^{rl}_{tr,n}$ in column 5.

By comparing column 4 and column 2, we can see that using rendered training images from the target novel category can significantly boost the performance, which shows that \emph{our method is applicable to other novel categories as well}. Based on the results in column 3, we can see that using rendered images from another category (``human") can generally improve the performance but the performance gain is much smaller and even negative (\emph{e.g.}, ``cake"), compared with the performance gain (column 4 \emph{v.s.} column 2) using rendered images from the target novel category, which proves that  \emph{within-category cross-domain transfer is more effective than cross-category cross-domain transfer.}
Despite the performance gap between column 4 and the upper bound (column 5), our method has shown great potential to bridge the gap between real images and rendered images on all 5 novel categories.

We also show the harmonization results of both iDIH backbone and our CharmNet in Figure \ref{fig:novel_five_res}, which shows that our method can generate more visually appealing results closer to the ground-truth after utilizing the rendered images from the  target novel category.

\section{Limitations}\label{sec:limit}

\begin{figure}[t]
\centering
\includegraphics[width=0.6\linewidth]{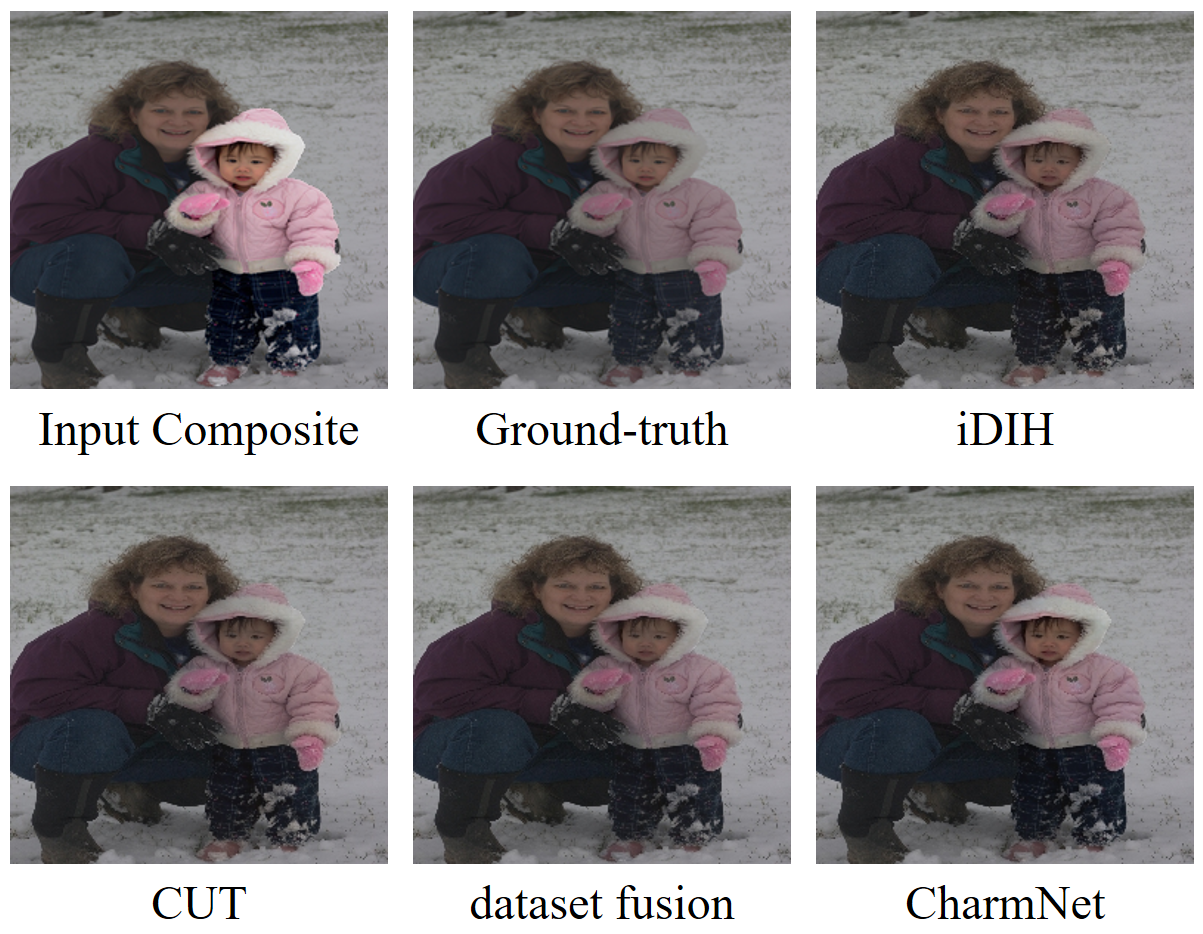}
   \caption{A sample of failure case of our CharmNet.}
\label{fig:failure}
\end{figure}

In the experiment, we observe that our CharmNet might encounter failures with the composite foreground is overexposed. For example, in Figure \ref{fig:failure}, the brim in the input composite image is overexposed. After harmonization, the result of our CharmNet may still be kind-of overexposed.

